\pdfoutput=1
\documentclass[10pt,twocolumn,letterpaper]{article}
\usepackage{iccv}

\usepackage{times}
\usepackage{epsfig}
\usepackage{graphicx}
\usepackage{amsmath}
\usepackage{amssymb}
\usepackage{amsmath}
\usepackage{amssymb}
\usepackage{multirow}
\usepackage{multicol}
\usepackage{bm}
\usepackage{array}
\usepackage{arydshln}
\usepackage{xcolor}
\usepackage{blindtext}
\usepackage{bm}
\usepackage{efbox}
\usepackage{amsthm}
\usepackage[english]{babel}
\usepackage{float}
\usepackage{subcaption}

\usepackage{wrapfig}
\usepackage{authblk}
\usepackage{hyperref}

\iccvfinalcopy 

\def\iccvPaperID{3993} 

\ificcvfinal\fi
\usepackage[accsupp]{axessibility}
\begin{document}
\makeatletter
\renewcommand\AB@affilsepx{, \protect\Affilfont}
\makeatother
\title{Unaligned Image-to-Image Translation by Learning to Reweight}

\author[1]{Shaoan Xie}
\author[2]{Mingming Gong}
\author[3]{Yanwu Xu}
\author[1]{Kun Zhang}
\affil[1]{Carnegie Mellon University}
\affil[2]{
The University of Melbourne}
\affil[3]{ University of Pittsburgh}
\maketitle
\ificcvfinal\thispagestyle{empty}\fi

\begin{abstract}
   Unsupervised image-to-image translation aims at learning the mapping from the source to target domain without using paired images for training. An essential yet restrictive assumption for unsupervised image translation is that the two domains are aligned, e.g., for the selfie2anime task, the anime (selfie) domain must contain only anime (selfie) face images that can be translated to some images in the other domain. Collecting aligned domains can be laborious and needs lots of attention.
   In this paper, we consider the task of image translation between two unaligned domains, which may arise for various possible reasons. 
   To solve this problem, we propose to select images based on importance reweighting and develop a method to  learn the weights and perform translation simultaneously and automatically. We compare the proposed method with state-of-the-art image translation approaches and present qualitative and quantitative results on different tasks with unaligned domains. Extensive empirical evidence demonstrates the usefulness of the proposed problem formulation and the  superiority of our method.

\end{abstract}

\section{Introduction}

In recent years, Image-to-Image (I2I) translation has been achieving remarkable success in transferring complex appearance changes across domains \cite{zhu2017unpaired, liu2017unsupervised}. In addition, many related tasks could also be formulated as I2I problems such as image super-resolution \cite{yuan2018unsupervised, chen2020unsupervised} and domain adaptation \cite{hoffman2018cycada, murez2018image}.

In supervised image translation, we are given paired data from source and target domains. Pix2pix \cite{isola2017image} applies conditional Generative Adversarial Network \cite{goodfellow2014generative,mirza2014conditional} to map the the source images to the target domain while enforcing a L1 distance loss between translated images and target images. Pix2pix can generate a sharp target image with sufficient paired training data. However, paired data are very difficult to collect or even do not exist (e.g., Van Gogh's painting to real photos). 
In the absence of paired data, unsupervised I2I translation methods have achieved impressive performance by combining GAN with proper constraints, such as cycle consistency \cite{zhu2017unpaired} and shared latent space assumption \cite{liu2017unsupervised}. 

\begin{figure}
    \centering
    \includegraphics[scale=0.45]{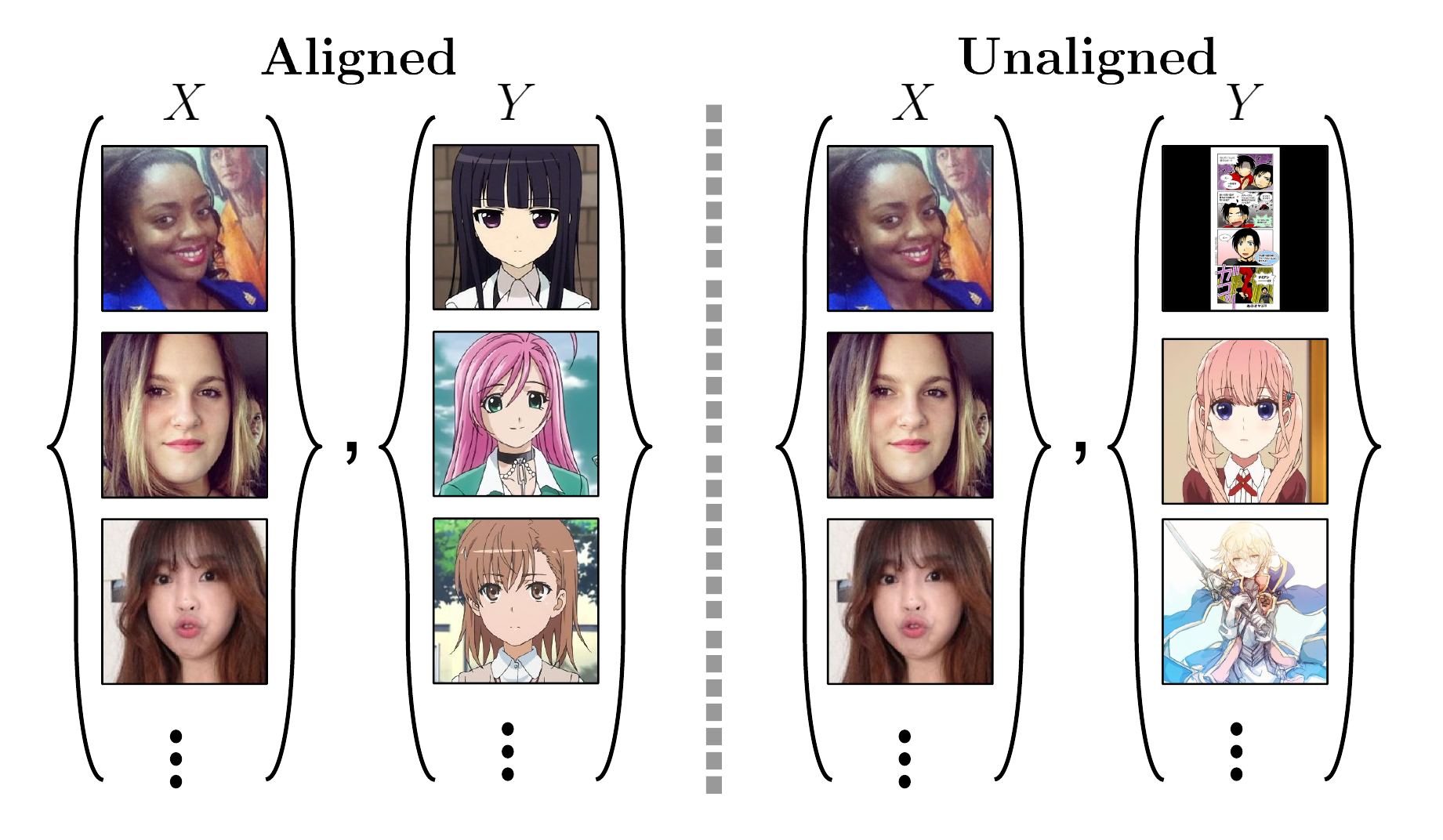}
    \caption{Example of aligned and unaligned domains. Left: selfie images as domain $X$ and anime face images as domain $Y$. Images in two domains are carefully selected and processed. Right: many unwanted anime images may appear in the domain $Y$ for many possible reasons, e.g., lack of human supervision.}
    \label{fig:illu}
    \vspace{-0.4cm}
\end{figure}
An essential assumption of unsupervised image translation is then that the domains used for training are {\it aligned}, which means that each image in one domain can be translated to some image in the other domain in a meaningful manner; in other words, there is some underlying relationship between the domains \cite{zhu2017unpaired}. For example, each of the two domains in the selfie2anime task include only female face images (Figure \ref{fig:illu}, left) of a similar style. 

However, collecting images for two domains which are guaranteed to be aligned needs a lot of attention. For instance, to collect the anime domain, Kim et al. \cite{kim2019u} first constructed an initial dataset consisting of 69,926 anime character images. Then they applied pre-trained anime face detector to extract 27,073 face images and then manually selected 3500 female face images as the training set. To collect the animal face dataset, Liu et al. \cite{liu2019few} manually labeled bounding boxes of 10,000 carnivorous animal faces in the images and selected images with high detection scores from Imagenet \cite{deng2009imagenet}.

To save efforts, one may consider the setting with unaligned domains since they are much cheaper to obtain. For example, to obtain the anime face domain, we may also apply the anime face detector as Kim et al. did, and then just treat the detected results as images in the desired domain. Without any human supervision, the constructed domain may contain many unwanted anime images, e.g., anime body or even anime book images as shown in Figure \ref{fig:illu} (right). These unaligned images may harm the image translation quality and can even cause the failure of some image translation methods (e.g., see Figure \ref{fig:comp_syn}).      

We therefore seek an algorithm that can learn to translate between {\it unaligned} domains where some images in either domain may be unrelated to the main task (Figure \ref{fig:illu}, right) and thus should not be considered for translation. For brevity, we denote these images as unaligned images. We further assume that there are unknown, aligned subsets $X_{a}\subseteq X$ and $Y_{a}\subseteq Y$, and our task is to discover such unknown subsets automatically and simultaneously learn the mapping between them. Inferring the unknown subsets $X_{a}$ and $Y_{a}$ seems to be challenging since we are given only two unaligned domains $X$ and $Y$. To address this issue, we propose to reweight (or ``select") each sample with importance $\beta$ during the adversarial distribution matching process. Ideally, if $\beta$ is almost 0,  then the image is not in the aligned subset and hence not considered for translation.

Thus the problem boils down to learning appropriate importance weight for each sample for the purpose of sensible translation. 
To address the importance weight estimation problem, we analyze the causal generating process of images and hypothesize that \emph{images in $X_a$ and $Y_a$ can be translated to the other domain faster than images in unaligned subsets since $X_a$ and $Y_a$ share the same content category}. Then we propose the reweighted adversarial loss which enables us to approximate the density ratios as well as performing image translation between two unknown aligned subsets $X_a, Y_a$. In addition, we also propose an effective sample size loss to avoid importance networks giving trivial solutions.
We apply the proposed method to various image-to-image translation problems and the large improvements over strong baselines on unaligned dataset demonstrate the efficacy of our proposed translation method as well as the validity of our hypothesis.  Code and data are available
at \href{https://github.com/Mid-Push/IrwGAN}{\textcolor{blue}{https://github.com/Mid-Push/IrwGAN}}.

\section{Related Works}
\label{sec:relatedwork}

\textbf{Image Translation}
Contemporary image-to-image translation approaches leverage the strong power of Generative Adversarial Network (GAN) \cite{goodfellow2014generative} to generate high-fidelity images. Paired image-to-image translation methods adopt reconstruction loss between the result and target to preserve the content of the input image \cite{isola2017image}. In contrast, there is no paired data available in the task of unsupervised image translation. To address this issue, cycle consistency is proposed to reduce the number of possible mappings in the function space. It enforces a one-to-one mapping between source and target domains  and is shown to achieve impressive visual performance \cite{kim2017learning, zhu2017unpaired, yi2017dualgan}; however, the one-to-one correspondence may not be enough to preserve content and many methods are proposed to facilitate better image translation \cite{tang2019attention, kim2019u, mejjati2018unsupervised, wu2019transgaga}. Alternatively, shared latent space assumption \cite{huang2018multimodal, liu2018unified, lee2018diverse,  liu2017unsupervised, liu2019few} and relationship preservation \cite{benaim2017one, zhang2019harmonic, fu2019geometry, park2020contrastive, amodio2019travelgan} also demonstrated their efficacy in image translation.
Recently multi-modal and multi-domain translation are gaining wide popularity \cite{mo2018instagan,ma2018exemplar,cao2019multi,huang2018multimodal, liu2018unified, lee2018diverse, yu2019multi,choi2018stargan, zhu2017toward, almahairi2018augmented, shen2019towards, nizan2020breaking}. However, unlike
the above prior works, we aim to learn the mapping with unaligned domains.

\textbf{Importance Reweighting} Importance reweighting is an important technique in various fields, including domain adaptation \cite{huang2007correcting,zhang2013domain,gretton2009covariate, sugiyama2008direct, yu2020label, gong2016domain, yan2017mind} and label-noise learning \cite{liu2015classification, xia2019anchor, zhang2020domain}. Their settings are very different from the problem we aim to solve. We are not given two sets of data points from which the density ratio is to be estimated.  We use importance reweighting as a way to learn to select proper images for translation. Furthermore, in our task, we need to reweight samples in two domains simultaneously. This treatment, together with the property that aligned image subsets are easier to be translated to each other, helps achieve automated image selection and translation. Without this property, the problem to be solve might be ill-posed. 
There are also some importance reweighting applications on generative models \cite{wu2016importance, tao2018chi, grover2019bias, diesendruck2019importance, wu2020improving, che2017maximum, hu2017unifying, hjelm2017boundary}. In this line of research, \cite{wu2020improving} reweights the fake samples of the generator by the exponential score of discriminator output to help discriminator training. \cite{hu2017unifying} proposes to assign normalized discriminator score for samples to achieve a tighter lower bound for GAN following the importance reweighting Variational autoencoder \cite{burda2015importance}. \cite{che2017maximum} proposes to  reweight each sample with the normalized ratio of discriminator output on a delayed copy of the generator to stabilize the training of GAN.  
These models \cite{hu2017unifying,wu2020improving,che2017maximum} apply known statistics to reweight the samples for better GAN training. In contrast, our importance weight is unknown and our goal is to learn such importance weight.


\section{Image Translation with Unaligned Domains}
Given collected yet unaligned domains $X$ and $Y$ with training samples $\{x_{i}\}_{i=1}^{N} \in X$ and $\{y_{j}\}_{j=1}^{M}\in Y$, our goal is to translate proper images from one domain to the other domain while
alleviating the detrimental effects brought by the unaligned images in two domains. We denote the aligned subsets of two domains by $X_a$ and $Y_a$, respectively. On the other hand, $X_u=X\setminus X_a$ and  $Y_u=Y\setminus Y_a$ correspond to the subsets that are not aligned in two domains. 

Our model includes two mappings $G: X\rightarrow Y$ and
$F: Y\rightarrow X$ and two discriminators $D_Y$ and $D_X$. In addition, we introduce two importance weight networks $\beta_X$ and
$\beta_Y$ , where $\beta_X$ is the weight applied on $\{x_i\}$ and $\beta_Y$
 the weight on $\{y_j\}$. Naturally, our objective includes
four types of terms: a reweighted adversarial loss to learn importance weights and perform image translation, an effective sample size loss to control
how many images are selected (in a soft manner) by importance weights, a reweighted
cycle consistency loss to keep the one-to-one mapping between two domains and a reweighted identity loss to keep networks conservative. For
brevity, we refer to our method as Importance Reweighting Generative Adversarial Network (\texttt{IrwGAN}).

\subsection{Connection between Aligned and Unaligned Translation}
\label{subsec:learnirw}
If the given domains $X$ and $Y$ are aligned (which is not the case in our setting), to transfer images from $X$ to $Y$, one may apply least squares adversarial loss \cite{mao2017least} directly to match the distributions between $P_{Y}$ and $P_{G(X)}$ with the mapping function $G:X\rightarrow Y$ and its discriminator $D_{Y}$ by solving: 
\begin{small}
\begin{align}
\label{eq:base}
    \min_{G}&\max_{D_{Y}}  ~~~~ \mathcal{L}_{\text{GAN}}(X,Y) \\
    =& \mathbb{E}_{x\sim P_{X}} [(1-D_{Y}(G(x)))^{2}]+ \mathbb{E}_{y\sim P_{Y}} [D_{Y}(y)^{2}].\nonumber
\end{align}
\end{small}
\vspace{-0.4cm}

However, $X$ and $Y$ are unaligned. Matching the unaligned images can harm the translation performance and we only want to match the distributions of unknown aligned subsets, In other words, we would like to optimize $\mathcal{L}_{\text{GAN}}(X_a,Y_a)$. 
We can observe that
\vspace{-0.2cm}

\begin{align}
 \label{eq:irw}
   &\mathcal{L}_{\text{GAN}}(X_a,Y_a)  \\
    = & \mathbb{E}_{x\sim P_{X_{a}}} [(1-D_{Y}(G(x)))^{2}] + \mathbb{E}_{y\sim P_{Y_{a}}}[D_{Y}(y)^{2}] \nonumber\\
    = & { \int_{x} {\scriptstyle P_{X}(x) \frac{P_{X_{a}}(x)}{P_{X}(x)}} [(1-D_{Y}(G(x)))^{2}]  dx} + \nonumber \\ 
    &~~~~~~~~~~~~~~~~~~~~~~~~~~~~~~ { \int_{y} {\scriptstyle P_{Y}(y) \frac{P_{Y_{a}}(y)}{P_{Y}(y)} } [D_{Y}(y)^{2}] dy}\nonumber\\ 
    = & \mathbb{E}_{x\sim P_{X}} {\scriptstyle \bm{\frac{P_{X_{a}}(x)}{P_{X}(x)}}} [(1-D_{Y}(G(x)))^{2}] + \nonumber\\
    &~~~~~~~~~~~~~~~~~~~~~~~~~~~~~~ \mathbb{E}_{y\sim P_{Y}}{\scriptstyle \bm{\frac{P_{Y_{a}}(y)}{P_{Y}(y)}}}  [D_{Y}(y)^{2}] .\nonumber
\end{align}
It shows that even though the given domains $X,Y$ are unaligned, if we reweight each sample $x\in X$  and $y\in Y$ with the corresponding density ratio $\frac{P_{X_{a}}(x)}{P_{X}(x)}$ and $\frac{P_{Y_{a}}(y)}{P_{Y}(y)}$, respectively, we are actually optimizing $\mathcal{L}_{\text{GAN}}(X_a, Y_a)$.

However, the density ratios are unknown. In light of this observation, we apply two networks $\beta_X, \beta_Y$ and use their outputs to reweight each sample. If we are able to learn $\beta_X(x)\approx \frac{P_{X_{a}}(x)}{P_{X}(x)}, \beta_Y(y) \approx \frac{P_{Y_{a}}(y)}{P_{Y}(y)}$, optimizing the generators $G,F$ and discriminators $D_Y, D_X$ is equivalent to performing image translation between unknown aligned subsets $X_a, Y_a$. 



\subsection{Learning to Select Images and Translate}


As shown above, we aim to find the density ratios with our importance networks $\beta_X$ and $\beta_Y$, without having access to aligned subsets. However, although we can restrict the outputs of $\beta_X$ and $\beta_Y$ to be valid density ratios, there can still exist many unwanted solutions. For example, images not in the aligned subsets may be translated to each other with neural networks of high  capacity and long enough training; then the estimated importance weights could be just one for all images, failing to select the correct aligned subset. Hence, we need to formulate and exploit proper constraints on the property of aligned subsets to find meaningful solutions of $\beta_X,\beta_Y$ and achieve successful translation.

To this end, we formulate the following \emph{quicker translation hypothesis}:  \emph{images in $X_a$ and $Y_a$ can be translated to the other domain faster than images in $X_u$ and $Y_u$}. With this hypothesis, the images in the aligned set will first be selected to match in the two domains, while the images in the unaligned subsets will have a relatively large adversarial loss and hence achieve very small importance weights. 
More specifically, we propose the following \emph{reweighted adversarial loss} to estimate the density ratios as well as matching the distributions:
\begin{align}
\label{eq:gan_loss}
   \min_{G,\beta_{X}} \max_{D_{Y}} &~~~~\mathcal{L}_{\text{GAN}}(X, Y) \\
   &= \mathbb{E}_{x\sim P_{X}} \beta_{X}(x)[(1-D_{Y}(G(x)))^{2}] +\nonumber\\
    &~~~~~~~~~\mathbb{E}_{y\sim P_{Y}}\beta_{Y}(y)[D_{Y}(y)^{2}], \nonumber
\end{align}
where $\beta_X(x)$ and $\beta_Y(y)$ represent the importance weights assigned to each sample. We introduce a similar reweighted loss for the mapping function $F: Y\rightarrow X$, i.e., $\mathcal{L}_{\text{GAN}}(Y,X)$. Note that $\beta_Y$ is not optimized in this loss function.



In each iteration, after we update the discriminator, we minimize the loss function $\mathcal{L}_{\text{GAN}}(X, Y)$, given in Equation (\ref{eq:gan_loss}), over $\beta_X$ and $G$. Intuitively, according to our hypothesis, images $x_a\in X_a$ are translated to the domain $Y$ faster, and thus the discriminator will assign a higher score to $G(x_a)$ than $G(x_u)$. Then $[(1-D_{Y}(G(x_a)))^{2}] \leq [(1-D_{Y}(G(x_u)))^{2}]$. If we assume that higher loss implies decreasing faster with the existing SGD optimization algorithms, then a consequence is that $\beta_X(x_u)$, compared to $\beta_X(x_a)$, will be decreased with  a larger rate proportional to $[(1-D_{Y}(G(x_u)))^{2}]$ as the optimization procedure  (\ref{eq:gan_loss}) proceeds.  $\beta_X(x_u)$ will then become smaller and finally are expected to be close to 0. In the experiment section 5.5, we visualize the learned importance weights and observe that the importance weights for unaligned images are close to 0.


\subsection{Analysis of the Hypothesis}

Although the quicker translation hypothesis is intuitive, one may wonder how sensible it is and how it helps in image selection for translation.  
In this section, we will look into the generating process of images and provide an  analysis of the hypothesis. Illustrative experiments are also provided to support our hypothesis.

\begin{wrapfigure}{r}{0.15\textwidth}
 \vspace{-4mm}
   \begin{center}
     \includegraphics[width=0.15\textwidth]{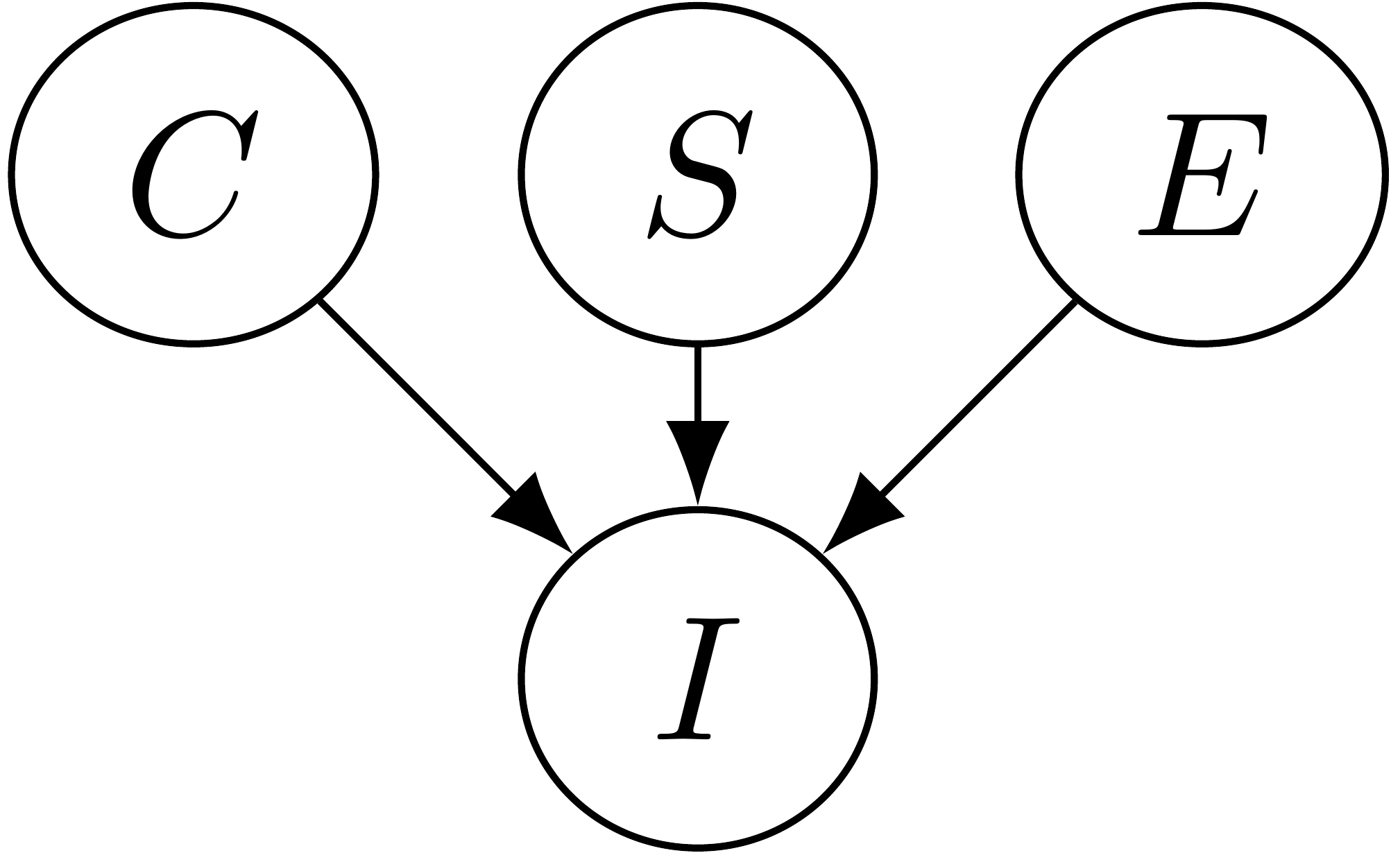}
   \end{center}
   \vspace{-4mm}
   \caption{Causal generating process of images.}
   \label{fig:causal_graph}\vspace{-2mm}
 \end{wrapfigure}

Similar to the partial shared latent space assumption in \cite{huang2018multimodal, lee2018diverse}, we assume that \emph{content category} $C$, \emph{style} $S$ and \emph{random input} $E$ (causally) generate the final image $I$, as illustrated in Figure \ref{fig:causal_graph}. Naturally, images in aligned subsets $X_a, Y_a$ are expected to share the same content category and differ only in the influence of style. Let us consider the same content category, denoted by $C_1$, which corresponds to the aligned subsets. Corresponding images are generated according to $I = F_{C_1,S}(E)$, where $F$ is the generation function, $S$ is the style indicator, and $E$ is the random input. ($E$ introduces the randomness in the images with the same content category and the same style, which is expected to remain the same during translation.)

The causal process is $C\rightarrow I$, and the two domains have different styles. The minimal change principle of the causal systems \cite{huang2020causal}, as well as the quicker adaptation method for learning causality from data with changing distributions \cite{bengio2019meta} suggests that $F_{C1,S2}$ is ``close" to $F_{C1,S1}$; it is easy to adapt one to the other, given that the cause $C_1$ does not change. As a consequence, the translation function from style $S_1$ to $S_2$, which can be written as $I_Y = F_{C1,S2} (F^{-1}_{C1,S2}(I_X))$, is rather simple and easy to learn as well.

If we also change the content category from $C_1$ and $C_2$, in addition to the change in style, then the (unaligned) image subsets are generated by functions $F_{C_1, S_1}$ and $F_{C_2,S_2}$ in the source and target domains, respectively. The two functions may be greatly different because of the additional change in the content category. The
(unnatural) translation function, which can be written as $F_{C_2,S_2} (F^{-1}_{C_1,S_2}(I_X))$, is expected to be more complex and difficult to learn than that corresponds to the same content category.

\begin{figure}
 
    \centering
    \includegraphics[scale=0.5]{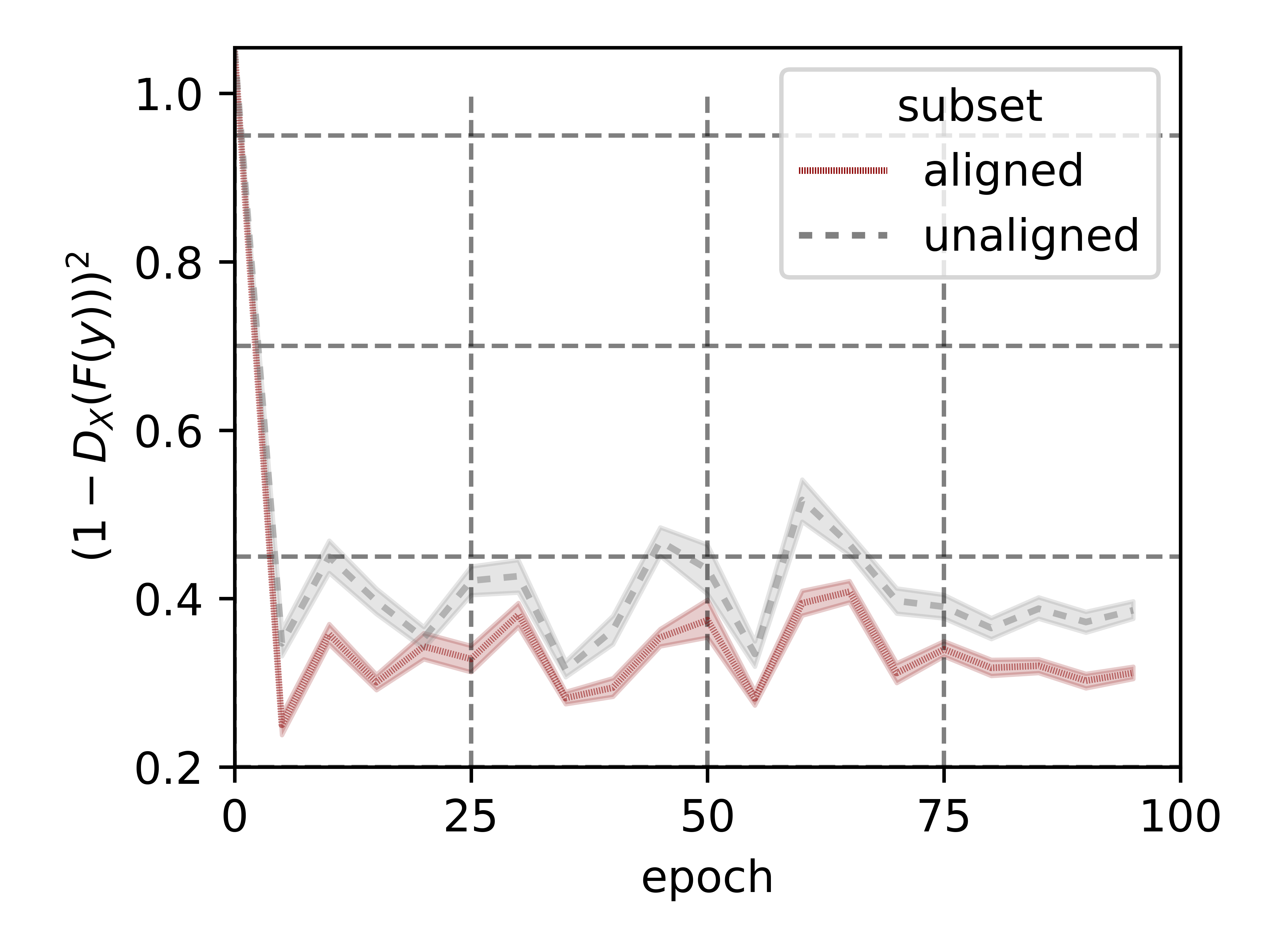}
\vspace{-0.4cm}
\caption{The trend of $(1-D_X(F(y)))^2$ for the task $A_U\rightarrow S$ where $X=S$ is the selfie face domain and $Y=A_U$ is the anime domain which contains anime faces (aligned subsets) and non-face anime images (unaligned subsets).}
\vspace{-0.5cm}
\label{fig:loss_trend}
\end{figure}

Now let us illustrate the hypothesis with real images. We use existing aligned face selfie2anime dataset as aligned subsets and we construct the unaligned subsets with non-face anime images. 
We apply CycleGAN \cite{zhu2017unpaired} on the unaligned dataset (We replace the discriminator with our improved discriminator (see section \ref{sec:impl})). 
As illustrated in Figure \ref{fig:loss_trend}, the term $(1-D_X(F(y)))^2$ is consistently larger for those unaligned images $Y_u$. An important consequence of our hypothesis is that  $(1-D_X(F(y)))^2$ should be larger for unaligned images. The empirical evidence well aligns with this consequence. For more detail of this illustration, please refer to the supplementary material.

\subsection{Effective Sample Size for Translation}
\label{subsec:ess}
\textbf{ Density ratio constraint}. Previous analysis suggests that $\beta_X$ and $\beta_Y$ should approximate the density ratios. Thus we regularize our importance network outputs such that they are valid density ratios. Below we discuss the constraints on $\beta_X$, which also apply to $\beta_Y$. The first constraint is that it has to be non-negative and the second constraint is that the $L_1$ norm is fixed.
The reason for the first constraint is obvious and as for the second constraint, we have
$\mathbb{E}_{x\sim P_{X}}\frac{P_{X_{a}}(x)}{P_{X}(x)}=\int P_{X}(x)\frac{P_{X_{a}}(x)}{P_{X}(x)}dx=1.$
 Thus we require the outputs of importance networks $\beta_{X},\beta_{Y}$ to have expectation 1. Empirically we need $\frac{1}{n}\sum_{i=1}^{n}\beta_{X}(x_{i})=1$ for a batch of images $\left\{x_{1},...,x_{n}\right\}$ where $n$ is the batch size. A same constraint also applies to $\beta_{Y}$. 

\textbf{Effective sample size loss }. 
The reweighted adversarial loss discards the unaligned images effectively through assigning them low importance weights. But there is no constraint on importance weights for aligned images. One may end up with the trivial case where networks assign low values to almost all images (including aligned images), that is, only few images are selected for translation.  This is also observed in section \ref{subsec:exp_lambda_nos}. 
To address this issue, we propose the \emph{effective sample size loss} to allow more aligned images being selected for translation:
    \begin{align} \vspace{-3mm}
        \min_{\beta_{X}} \mathcal{L}_{\text{ESS}}(X)=||\beta_{X}||_{2}.
        \label{eq:nos}\vspace{-3mm}
    \end{align}
A same loss $\mathcal{L}_{\text{ESS}}(Y)$ is introduced on the importance network $\beta_Y$.
To understand why minimizing the above function will maximize the size of the paired subset for translation, one can see that under only the fixed $L_1$ norm constraint on $\beta_X$,  the above term is minimized when $\beta_{X}(x)=1$ for every image $x$ in domain $X$, which means that all images are selected for translation. In contrast, if we only select one image for translation, i.e., $\beta_{X}(x_{i})=n$ for one image $x_{i}$ while $\beta_{X}(x_{j})=0$ for the remaining images $\{x_{j}\}$, $||\beta_{X}||_2$ reaches its maximum. Therefore, we can control the effective sample size by assigning different weights on the objective in Equation (\ref{eq:nos}). Our effective sample size loss was inspired by \cite{gretton2009covariate} which proves that the effective sample size in domain adaptation may be defined as $n/||\beta_{X}||^2$ under some conditions. Therefore, we encourage networks to select more samples by penalizing the $L_2$ norm of the importance weight vectors.

\subsection{Additional Regularizations}

Adversarial loss in Equation (\ref{eq:gan_loss}) can help match the distributions between $P_{Y_{a}}$ and $P_{G(X_{a})}$. However, 
with a large enough capacity, a network can map the same
set of input images to any random permutation of images in
the target domain \cite{zhu2017unpaired} and thus we need additional constraints to avoid it. To further regularize the mapping networks $G$ and $F$, we also apply the \emph{reweighted cycle consistency loss} to enforce a one-to-one mapping between the importance reweighted domain distributions:
\begin{align}
    \min_{G,F} \mathcal{L}_{\text{cyc}}(X,Y)&=\mathbb{E}_{x\sim P_{X}} \beta_{X}(x) \| x-F(G(x))\|_{1}\nonumber \\
    &~~~~~+\mathbb{E}_{y\sim P_{Y}} \beta_{Y}(y) \| y-G(F(y))\|_{1},
\end{align}
together with the \emph{reweighted identity loss} to keep networks conservative \cite{zhu2017unpaired}:
\begin{align}
    \min_{G,F} \mathcal{L}_{\text{idt}}(X,Y)&=\mathbb{E}_{x\sim P_{X}} \beta_{X}(x) \| x-F(x)\|_{1}\nonumber \\
    &~~~~~+\mathbb{E}_{y\sim P_{Y}} \beta_{Y}(y) \| y-G(y)\|_{1}.
\end{align}

\subsection{Full Objective}

Our full objective is 
\begin{align}
    \min_{G,F}\max_{D_{X},D_Y} 
    &~~~~\mathcal{L}_{\text{GAN}}(X,Y)  
    ~~+\mathcal{L}_{\text{GAN}}(Y, X)  \nonumber\\
    &~~~~~~~~+\lambda_{\text{cyc}} \mathcal{L}_{\text{cyc}}(X,Y)  +\lambda_{\text{idt}}\mathcal{L}_{\text{idt}}(X,Y), \nonumber
\end{align}
\vspace{-1.1cm}

\begin{align}
    \min_{\beta_{X}} \mathcal{L}_{\text{GAN}}(X, Y)   + \lambda_{\text{ESS}}\mathcal{L}_{\text{ESS}}(Y),  \nonumber
\end{align}
  \vspace{-1.1cm}

\begin{align}
\label{eq:fullobj}
    \min_{\beta_{Y}} \mathcal{L}_{\text{GAN}}(Y, X)  + \lambda_{\text{ESS}}\mathcal{L}_{\text{ESS}}({Y}),
\end{align}
where $\lambda_{\text{cyc}},\lambda_{\text{idt}}$, and $\lambda_{\text{ESS}}$ control the relative importance of different losses.
In addition, importance weight vectors $\beta_{X}$ and $\beta_{Y}$ need to be non-negative and have fixed $L_1$ norm, as addressed by parameterization in section \ref{sec:impl}.

\section{Implementation}
\label{sec:impl}


\textbf{Mapping and discriminator network architecture} We adopt the generator architecture used in CycleGAN \cite{zhu2017unpaired}, which contains 9 residual blocks \cite{he2016deep}. To capture the global structure and local region, we apply two discriminators for each mapping direction; one consists of 3 downsampling convolutional layers and the other consists of 5 downsampling convolutional layers. 

\textbf{Importance weight network architecture} The outputs have to satisfy the \emph{non-negativity} and \emph{fixed sum} constraints, and we address these two constraints by reparameterization: 
For the importance networks $\beta_{X}$ and $\beta_{Y}$, we firstly downsample images to 64$\times$64 in order to save memory. Then we apply 4 convolutional networks with kernel size 4, stride 2, padding 1. Then we append a fully connected network to the output and use a Softmax layer to normalize the outputs so that they sum to 1 and non-negative. Finally, we multiply the outputs by the batch size to make $\beta$ satisfy the fixed sum constraint.

\textbf{Training details} 
%
For all the experiments, we set $\lambda_{\text{idt}}=\lambda_{\text{cyc}}=10$ in Equation (\ref{eq:fullobj}). We find that $\lambda_{\text{ESS}}=1$ works well for all our experiments. We use the Adam solver \cite{kingma2014adam} with the learning rate of 0.0001. We train networks from scratch and keep the same learning rate for the first 50 epochs and linearly decay the rate to zero over the next 50 epochs. We use 10,000 images in each epoch. Since we need to input a batch of images into the network at the same time while the resolution of input images is high, e.g., 256$\times$256, we use the gradient accumulation trick to avoid GPU memory explosion. 
We set the batch size to 20 in all experiments.


\setlength{\tabcolsep}{0.2pt}
 \begin{figure*}[t]
 \centering
\begin{tabular}{c c c c c c c c c c}
{\footnotesize Task}&{\footnotesize Input} & {\footnotesize Cyclegan \cite{zhu2017unpaired}} & {\footnotesize MUNIT\cite{huang2018multimodal}} &{\footnotesize GcGAN \cite{fu2019geometry}}   &{\footnotesize NICE-GAN \cite{chen2020unsupervised}}   &{\footnotesize Baseline } & {\footnotesize \textbf{IrwGAN (Ours)}}  \\
{ $H_{S}\rightarrow Z_{D}$}&\multicolumn{1}{m{1.92cm}}{\frame{\includegraphics[width=1.63cm,height=1.63cm]{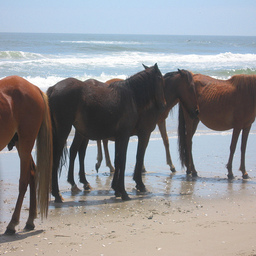}}}& 
\multicolumn{1}{m{1.92cm}}{ \frame{\includegraphics[width=1.63cm,height=1.63cm]{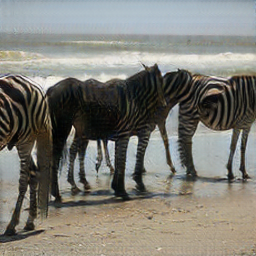}}}&
 \multicolumn{1}{m{1.92cm}}{\frame{\includegraphics[width=1.63cm,height=1.63cm]{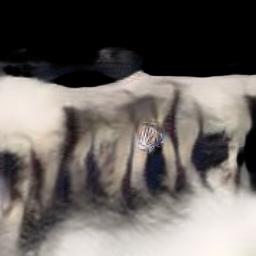}}}&
\multicolumn{1}{m{1.92cm}}{ \frame{\includegraphics[width=1.63cm,height=1.63cm]{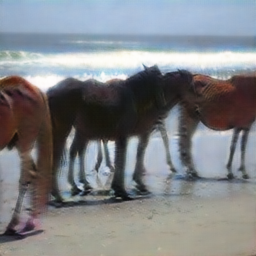}}}&
 \multicolumn{1}{m{1.92cm}}{ \frame{\includegraphics[width=1.63cm,height=1.63cm]{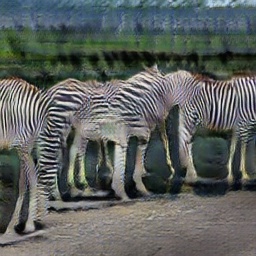}}}&
\multicolumn{1}{m{1.92cm}}{  \frame{\includegraphics[width=1.63cm,height=1.63cm]{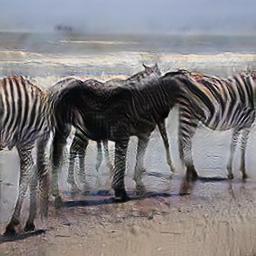}}}&
 \multicolumn{1}{m{1.92cm}}{ \frame{\includegraphics[width=1.63cm,height=1.63cm]{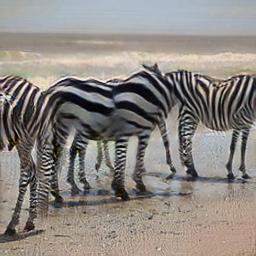}}}
   \\
  { $Z_{D}\rightarrow H_{S}$}&\multicolumn{1}{m{1.92cm}}{ \frame{\includegraphics[width=1.63cm,height=1.63cm]{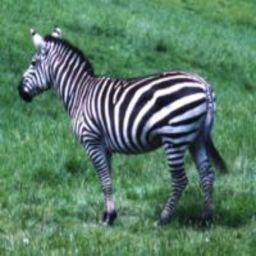}}}& 
  \multicolumn{1}{m{1.92cm}}{\frame{\includegraphics[width=1.63cm,height=1.63cm]{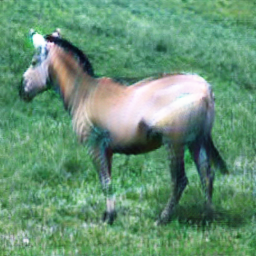}}}&
  \multicolumn{1}{m{1.92cm}}{\frame{\includegraphics[width=1.63cm,height=1.63cm]{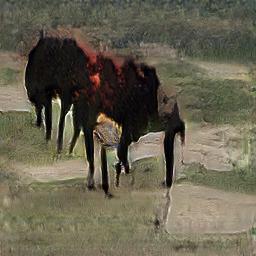}}}&
 \multicolumn{1}{m{1.92cm}}{ \frame{\includegraphics[width=1.63cm,height=1.63cm]{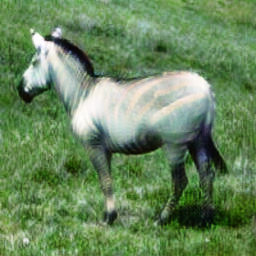}}}&
   \multicolumn{1}{m{1.92cm}}{\frame{\includegraphics[width=1.63cm,height=1.63cm]{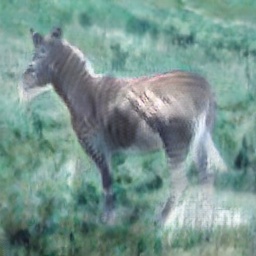}}}&
   \multicolumn{1}{m{1.92cm}}{\frame{\includegraphics[width=1.63cm,height=1.63cm]{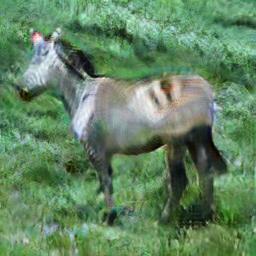}}}&
 \multicolumn{1}{m{1.92cm}}{ \frame{\includegraphics[width=1.63cm,height=1.63cm]{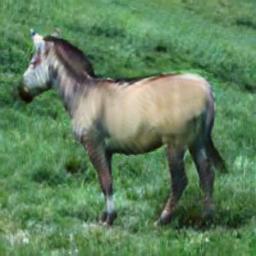}}}\\
 
 { $S\rightarrow A_{U}$}&\multicolumn{1}{m{1.92cm}}{\frame{\includegraphics[width=1.63cm,height=1.63cm]{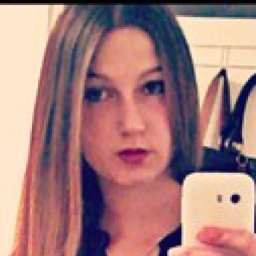}}}& 
 \multicolumn{1}{m{1.92cm}}{\frame{\includegraphics[width=1.63cm,height=1.63cm]{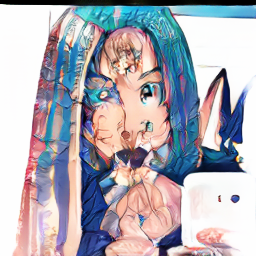}}}&
 \multicolumn{1}{m{1.92cm}}{\frame{\includegraphics[width=1.63cm,height=1.63cm]{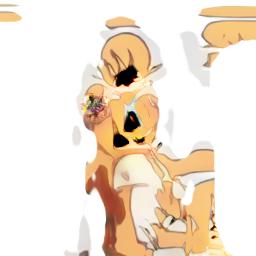}}}&
 \multicolumn{1}{m{1.92cm}}{\frame{\includegraphics[width=1.63cm,height=1.63cm]{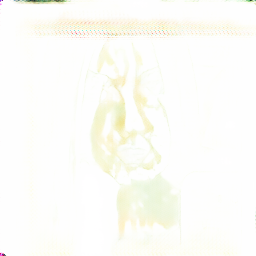}}}&
 \multicolumn{1}{m{1.92cm}}{ \frame{\includegraphics[width=1.63cm,height=1.63cm]{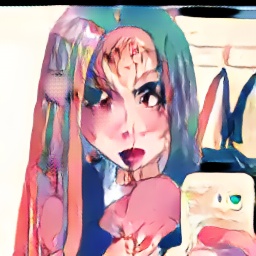}}}&
 \multicolumn{1}{m{1.92cm}}{ \frame{\includegraphics[width=1.63cm,height=1.63cm]{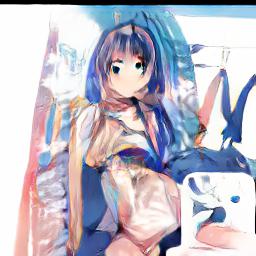}}}&
 \multicolumn{1}{m{1.92cm}}{ \frame{\includegraphics[width=1.63cm,height=1.63cm]{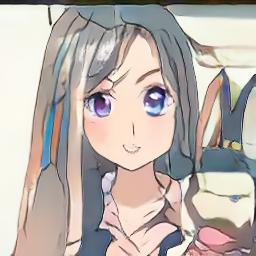}}}&
   \\
 { $A_{U}\rightarrow S$} &\multicolumn{1}{m{1.92cm}}{\frame{\includegraphics[width=1.63cm,height=1.63cm]{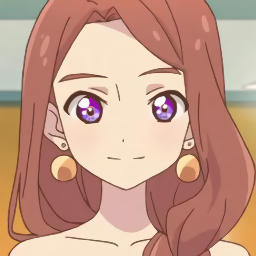}}}& 
 \multicolumn{1}{m{1.92cm}}{\frame{\includegraphics[width=1.63cm,height=1.63cm]{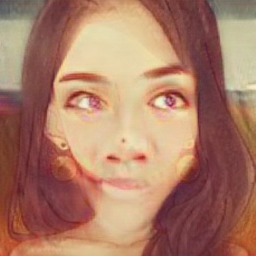}}}&
 \multicolumn{1}{m{1.92cm}}{\frame{\includegraphics[width=1.63cm,height=1.63cm]{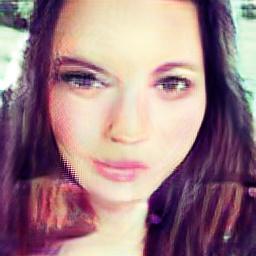}}}&
 \multicolumn{1}{m{1.92cm}}{\frame{\includegraphics[width=1.63cm,height=1.63cm]{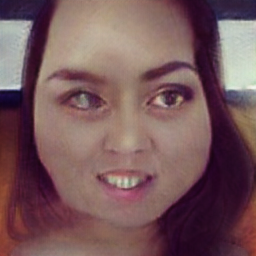}}}&
  \multicolumn{1}{m{1.92cm}}{\frame{\includegraphics[width=1.63cm,height=1.63cm]{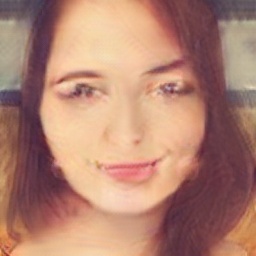}}}&
  \multicolumn{1}{m{1.92cm}}{\frame{\includegraphics[width=1.63cm,height=1.63cm]{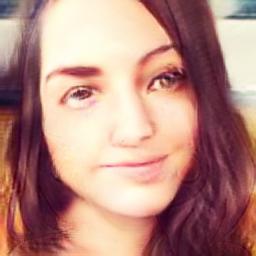}}}&
  \multicolumn{1}{m{1.92cm}}{\frame{\includegraphics[width=1.63cm,height=1.63cm]{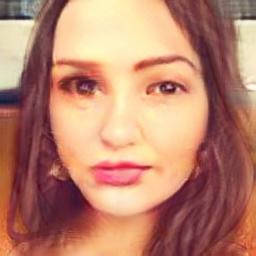}}}&
   \\
  { $C_{H}\rightarrow D_{A}$} &  \multicolumn{1}{m{1.92cm}}{\frame{\includegraphics[width=1.63cm,height=1.63cm]{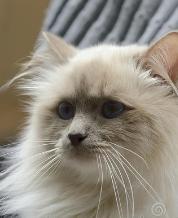}}}& 
 \multicolumn{1}{m{1.92cm}}{\frame{\includegraphics[width=1.63cm,height=1.63cm]{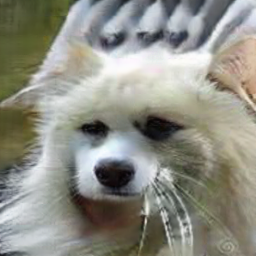}}}&
 \multicolumn{1}{m{1.92cm}}{\frame{\includegraphics[width=1.63cm,height=1.63cm]{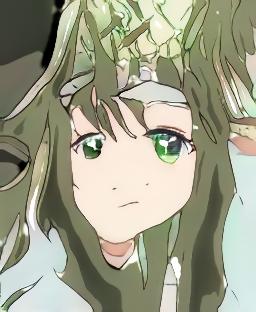}}}&
 \multicolumn{1}{m{1.92cm}}{\frame{\includegraphics[width=1.63cm,height=1.63cm]{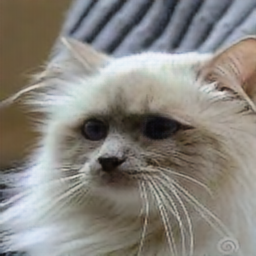}}}&
  \multicolumn{1}{m{1.92cm}}{\frame{\includegraphics[width=1.63cm,height=1.63cm]{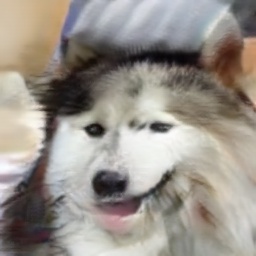}}}&
  \multicolumn{1}{m{1.92cm}}{\frame{\includegraphics[width=1.63cm,height=1.63cm]{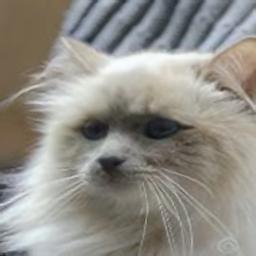}}}&
  \multicolumn{1}{m{1.92cm}}{\frame{\includegraphics[width=1.63cm,height=1.63cm]{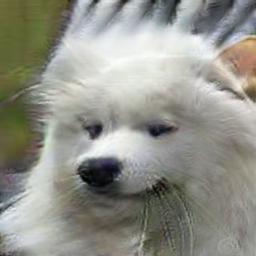}}}&
   \\
   
  { $D_{A}\rightarrow C_{H}$} &   \multicolumn{1}{m{1.92cm}}{\frame{\includegraphics[width=1.63cm,height=1.63cm]{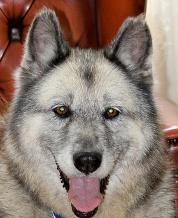}}}& 
 \multicolumn{1}{m{1.92cm}}{\frame{\includegraphics[width=1.63cm,height=1.63cm]{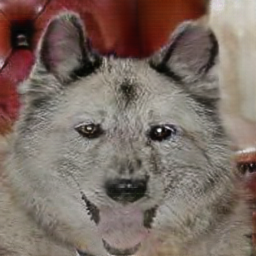}}}&
 \multicolumn{1}{m{1.92cm}}{\frame{\includegraphics[width=1.63cm,height=1.63cm]{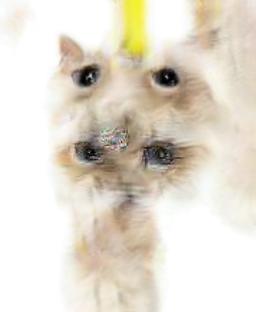}}}&
 \multicolumn{1}{m{1.92cm}}{\frame{\includegraphics[width=1.63cm,height=1.63cm]{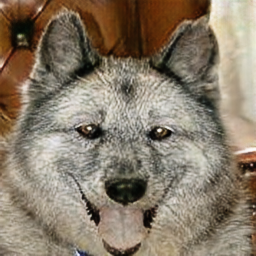}}}&
 \multicolumn{1}{m{1.92cm}}{\frame{ \includegraphics[width=1.63cm,height=1.63cm]{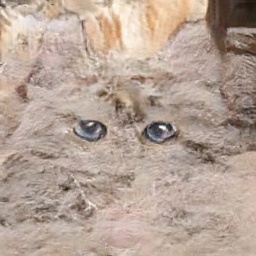}}}&
 \multicolumn{1}{m{1.92cm}}{\frame{ \includegraphics[width=1.63cm,height=1.63cm]{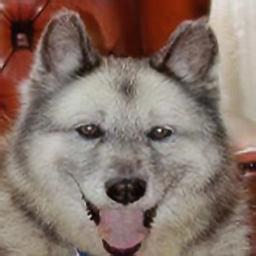}}}&
 \multicolumn{1}{m{1.92cm}}{\frame{ \includegraphics[width=1.63cm,height=1.63cm]{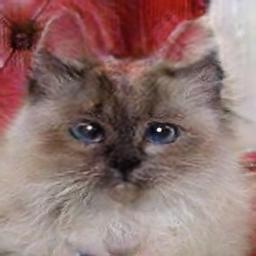}}}&
   \\   
   
   { $T_{B}\rightarrow L_{E}$} & 
       \multicolumn{1}{m{1.92cm}}{\frame{\includegraphics[width=1.63cm,height=1.63cm]{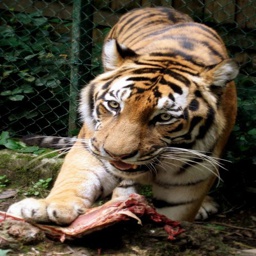}}}& 
 \multicolumn{1}{m{1.92cm}}{\frame{\includegraphics[width=1.63cm,height=1.63cm]{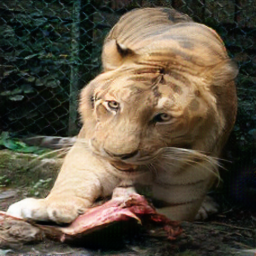}}}&
 \multicolumn{1}{m{1.92cm}}{\frame{\includegraphics[width=1.63cm,height=1.63cm]{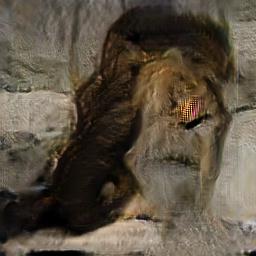}}}&
 \multicolumn{1}{m{1.92cm}}{\frame{\includegraphics[width=1.63cm,height=1.63cm]{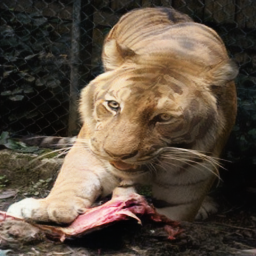}}}&
  \multicolumn{1}{m{1.92cm}}{\frame{\includegraphics[width=1.63cm,height=1.63cm]{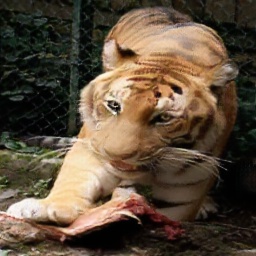}}}&
  \multicolumn{1}{m{1.92cm}}{\frame{\includegraphics[width=1.63cm,height=1.63cm]{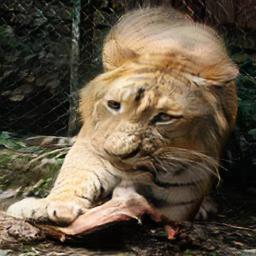}}}&
  \multicolumn{1}{m{1.92cm}}{\frame{\includegraphics[width=1.63cm,height=1.63cm]{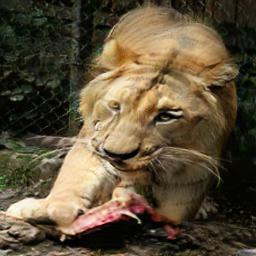}}}&
   \\{ $L_{E}\rightarrow T_{B}$} & 
       \multicolumn{1}{m{1.92cm}}{\frame{\includegraphics[width=1.63cm,height=1.63cm]{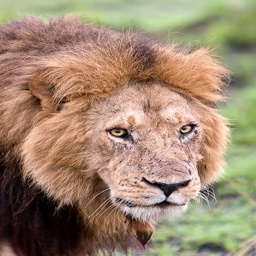}}}& 
 \multicolumn{1}{m{1.92cm}}{\frame{\includegraphics[width=1.63cm,height=1.63cm]{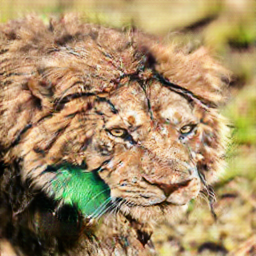}}}&
 \multicolumn{1}{m{1.92cm}}{\frame{\includegraphics[width=1.63cm,height=1.63cm]{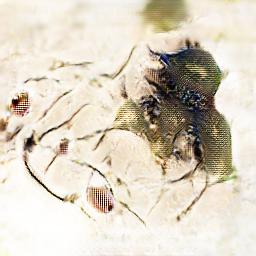}}}&
 \multicolumn{1}{m{1.92cm}}{\frame{\includegraphics[width=1.63cm,height=1.63cm]{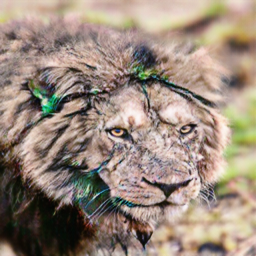}}}&
  \multicolumn{1}{m{1.92cm}}{\frame{\includegraphics[width=1.63cm,height=1.63cm]{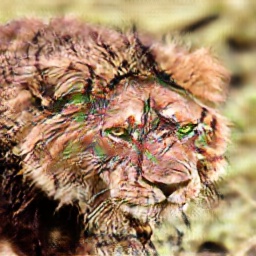}}}&
  \multicolumn{1}{m{1.92cm}}{\frame{\includegraphics[width=1.63cm,height=1.63cm]{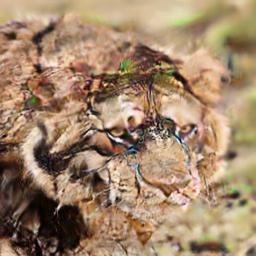}}}&
  \multicolumn{1}{m{1.92cm}}{\frame{\includegraphics[width=1.63cm,height=1.63cm]{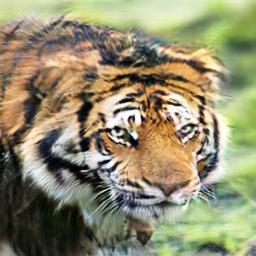}}}&
   \\
    \end{tabular}
\caption{: Visual comparisons of results by different algorithms (see the top row) on different datasets (given on the left). Abbreviations: (S)elfie, (A)nime, (H)orse, (Z)ebra, (C)at, (D)og, danb(U)roo, (T)iger, tiger (B)eetle, (L)ion, s(E)a lion, le(O)pard. $P_{Q}$ denotes the domain after add images in $Q$ to the original domain $P$.
}
\label{fig:comp_syn}
\vspace{-0.4cm}
\end{figure*}

\section{Experiments}
\label{sec:exp}

We first compare our approach against recent methods
for unsupervised image translation on different datasets. We then present the evaluation results of learned importance weights $\beta_{X},\beta_{Y}$. Finally, we investigate the effect of the proposed \emph{effective sample size loss} by varying  $\lambda_{\text{ESS}}$. 
\begin{table*}[ht]
    \centering
   
        \caption{The FID and KID ($\times$100) for different algorithms. Lower is better.  
        Abbreviations:(S)elfie, (A)nime, (H)orse, (Z)ebra, (C)at, (D)og, danb(U)roo, (T)iger, tiger (B)eetle, s(E)a lion. $P_{Q}$ denotes the unaligned domain after add images in $Q$ to the original domain $P$.  
        } \vspace{-2mm}
        \begin{small}
    \begin{tabular}{|c|p{1.6cm}|p{1.6cm}|p{1.6cm}|p{1.6cm}|p{1.6cm}|p{1.6cm}|p{1.6cm}|p{1.6cm}|}
    \hline
         \multirow{2}{*}{Method}&
         \multicolumn{2}{c|}{$H_{S}\rightarrow Z_{D}$} & \multicolumn{2}{c|}{$S\rightarrow A_{U}$}
         &\multicolumn{2}{c|}{$C_{H} \rightarrow D_{A}$} 
         &\multicolumn{2}{c|}{$T_{B}\rightarrow L_{E}$} \\ \cline{2-9} 
         &\hfil FID $\downarrow$&\hfil KID $\downarrow$&\hfil FID $\downarrow$&\hfil KID $\downarrow$&\hfil FID $\downarrow$&\hfil KID $\downarrow$&\hfil FID $\downarrow$&\hfil KID $\downarrow$
         \\
         \hline
         \centering
          CycleGAN \cite{zhu2017unpaired}& \hfil\hfil87.28& \hfil2.74 & \hfil100.43 & \hfil \textbf{1.98 }& \hfil68.97 & \hfil 2.03 & \hfil112.41 & \hfil 6.95  \\ \hline
          MUNIT \cite{huang2018multimodal}& \hfil287.79& \hfil22.19 & \hfil180.95 & \hfil 8.10& \hfil132.21 & \hfil 5.77 & \hfil335.52 & \hfil 25.35  \\\hline
           GcGAN \cite{fu2019geometry}& \hfil174.38 & \hfil 11.32 & \hfil267.73 & \hfil 20.92& \hfil 73.59 & \hfil 2.47 & \hfil110.76 & \hfil 6.48    \\\hline
           NICE-GAN \cite{chen2020reusing}& \hfil 166.61& \hfil3.79 & \hfil124.11 & \hfil 4.51 & \hfil229.00 & \hfil 18.80 & \hfil147.32 & \hfil 11.99  \\ \hline
         Baseline& \hfil106.84 & \hfil 4.01 & \hfil123.35 & \hfil 4.09 & \hfil64.79 & \hfil \textbf{1.89} & \hfil97.82 & \hfil 3.05  \\ \hline
         IrwGAN& \hfil\textbf{79.40} & \hfil \textbf{1.83 }& \hfil\textbf{93.75} & \hfil 2.58 & \hfil \textbf{60.97} & \hfil 2.07 & \hfil\textbf{84.91} & \hfil \textbf{2.34  } \\ 
         \hline
       
    \end{tabular}
    \vspace{0.2cm}
    
     \begin{tabular}{|c|p{1.6cm}|p{1.6cm}|p{1.6cm}|p{1.6cm}|p{1.6cm}|p{1.6cm}|p{1.6cm}|p{1.6cm}|}
    \hline
         \multirow{2}{*}{Method}&
         \multicolumn{2}{c|}{$Z_{D}\rightarrow H_{S}$} & \multicolumn{2}{c|}{$A_{U}\rightarrow S$}
         &\multicolumn{2}{c|}{$D_{A} \rightarrow C_{H}$}
         &\multicolumn{2}{c|}{$L_{E}\rightarrow T_{B}$}\\ \cline{2-9}
        &\hfil FID $\downarrow$&\hfil KID $\downarrow$&\hfil FID $\downarrow$&\hfil KID $\downarrow$&\hfil FID $\downarrow$&\hfil KID $\downarrow$&\hfil FID $\downarrow$&\hfil KID $\downarrow$
         \\
         \hline
          CycleGAN \cite{zhu2017unpaired}& \hfil 151.94& \hfil 4.63& \hfil 124.46 & \hfil  2.29& \hfil 96.94 & \hfil  3.43& \hfil 101.32 & \hfil  4.75  \\ \hline
          MUNIT \cite{huang2018multimodal}& \hfil 245.97& \hfil 10.59& \hfil 127.14 & \hfil  3.66& \hfil 174.32 & \hfil  7.11 & \hfil 304.80 & \hfil  26.33  \\\hline
           GcGAN \cite{fu2019geometry}& \hfil 161.75 & \hfil \textbf{3.45}& \hfil 133.58 & \hfil 3.77 & \hfil 153.83 & \hfil 8.71 & \hfil 130.91 & \hfil 8.83 \\\hline
           NICE-GAN \cite{chen2020reusing}& \hfil  166.54& \hfil {3.52}& \hfil 128.44 & \hfil  2.45 & \hfil 194.96 & \hfil  11.42 & \hfil 135.52 & \hfil  6.98  \\ \hline
         Baseline& \hfil 162.32 & \hfil  3.73 & \hfil \textbf{115.39} & \hfil  2.24 & \hfil 61.28 & \hfil  2.06 & \hfil 112.77 & \hfil  4.82\\ \hline
         IrwGAN& \hfil \textbf{142.98} & \hfil  3.74 & \hfil 119.86 & \hfil  \textbf{2.07 }& \hfil \textbf{53.46} & \hfil  \textbf{1.84 }& \hfil \textbf{77.47} & \hfil  \textbf{2.44 }  \\ 
         \hline
    \end{tabular}
    \end{small}
\label{tab:exp_fid_kid}
\end{table*}

\subsection{Dataset}
\label{subsec:dataset}

For simplicity, we use abbreviations for the datasets: (S)elfie, (A)nime, (H)orse, (Z)ebra, (C)at, (D)og, danb(U)roo, (T)iger, tiger (B)eetle, (L)ion and s(E)a lion. $P_{Q}$ denotes the domain after add images in $Q$ to the original domain $P$.

\vspace{-0.4cm}
\paragraph{$\bm{H_{S} \leftrightarrow  Z_{D}}$ and $\bm{C_{H}\leftrightarrow D_{A}}$.}
Because most of existing image translation datasets are carefully built to be aligned, to evaluate our method, we first construct two unaligned datasets using three image translation datasets: horse2zebra \cite{zhu2017unpaired} , selfie2anime \cite{kim2019u} , and cat2dog \cite{lee2018diverse} . For the constructed dataset ${H_{S} \leftrightarrow  Z_{D}}$, the main task is horse2zebra; we add selfie domain to the horse domain and we add dog domain to the zebra domain. 
For ${C_{H}\leftrightarrow D_{A}}$, the main task is cat2dog, we add horse domain to the cat domain and add anime domain to the dog domain.

\vspace{-0.4cm}
\paragraph{$\bm{S\leftrightarrow A_{U}}$.} To collect selfie2anime dataset, Kim et al. \cite{kim2019u} used the pretrained face detector to collect anime faces. It highly depends on the accuracy of the face detector. Therefore, it is interesting to consider the case where there is no pretrained face detector or the detector accuracy is low. To this end, we add 2869 anime images from the Danbooru anime dataset \cite{danbooru2019} to the anime face domain. The Danbooru dataset covers anime face, body, book, and many related images. 
Since only one domain is unaligned, we only learn $\beta_{Y}$ for this task.


\vspace{-0.4cm}
\paragraph{$\bm{T_{B}\leftrightarrow L_{E}}$.} We also consider a more realistic case where one uses search engines to obtain images: when searching for \emph{lion}, we may get not only images related to lion but sea lion images. In light of this observation, we use tiger class (1300 images) and tiger beetle class (1300 images) in Imagenet \cite{deng2009imagenet} as $T_{B}$ domain and lion class (1300 images) and sea lion class (1300 images) as $L_{E}$ domain. We select 100 tiger images and 100 lion images as the test set.


\subsection{Baselines and Metrics}
\label{subsec:baseline}

We compare our method with CycleGAN \cite{zhu2017unpaired}, MUNIT \cite{huang2018multimodal}, GcGAN \cite{fu2019geometry}, and NiceGAN \cite{chen2020reusing}. Different from CycleGAN, we use one global and local discriminator for each translation mapping and adopt different learning rate and batch size. To exclude possible effects by these differences and fully examine the effects of our proposed method, we run our method with $\beta_X(x)=\beta_Y(y)=1$ for all samples and we denote it as Baseline. 
For performance evaluation, we adopt the two commonly used metrics in image translation literature: the FID \cite{heusel2017gans} and KID score \cite{binkowski2018demystifying}. They measure the distribution divergence between the generated images and target images.



\begin{figure}
    \centering
    \includegraphics[scale=0.22]{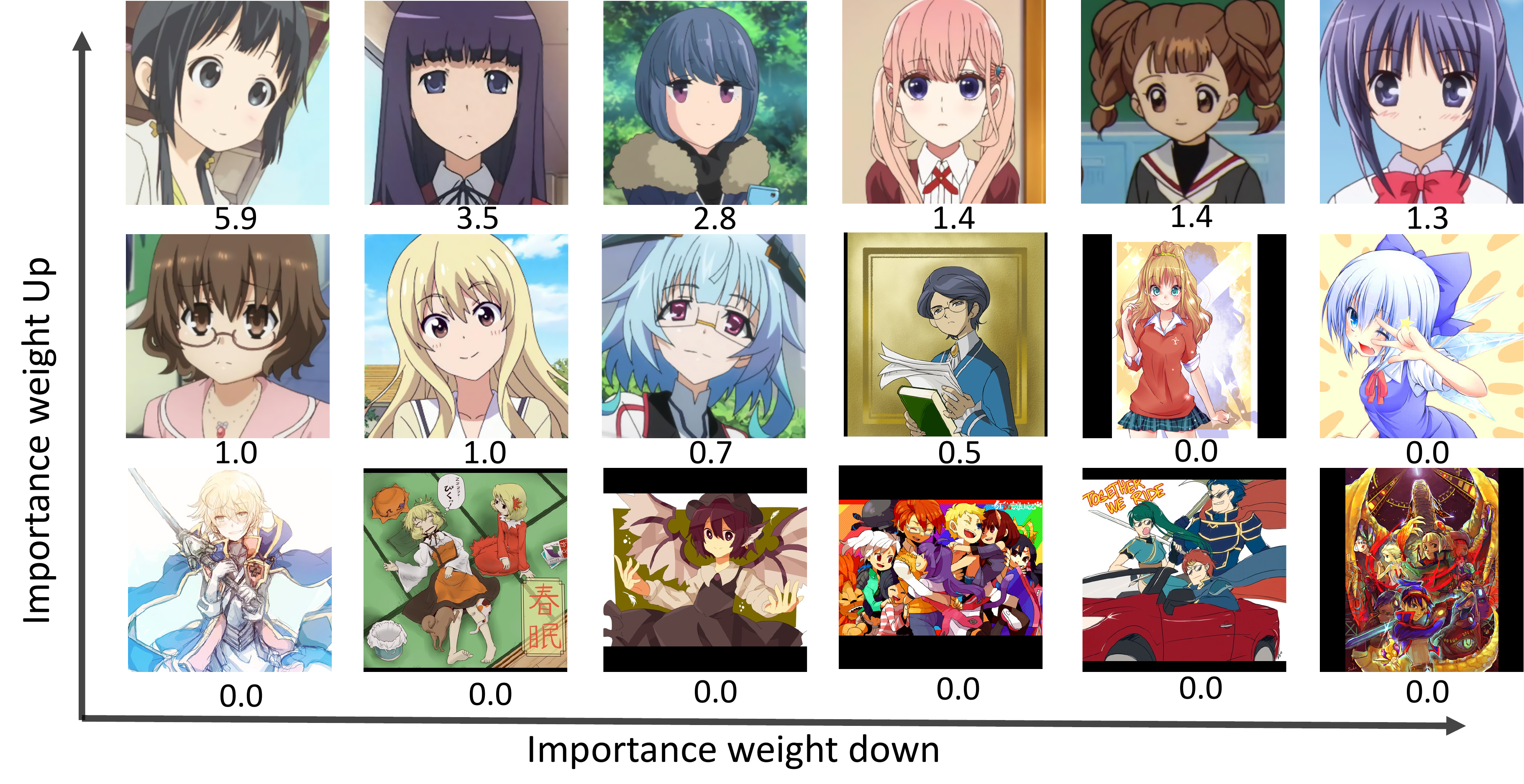} \vspace{-0.2cm}
    \caption{Examples of the learned importance weights for domain $A_{U}$ in the task $S\leftrightarrow A_{U}$.}
    \label{fig:vis_betas}
    \vspace{-0.8cm}
\end{figure}

\begin{figure*}[ht]
    \centering
   \begin{tabular}{ccccc}
    \includegraphics[scale=0.2]{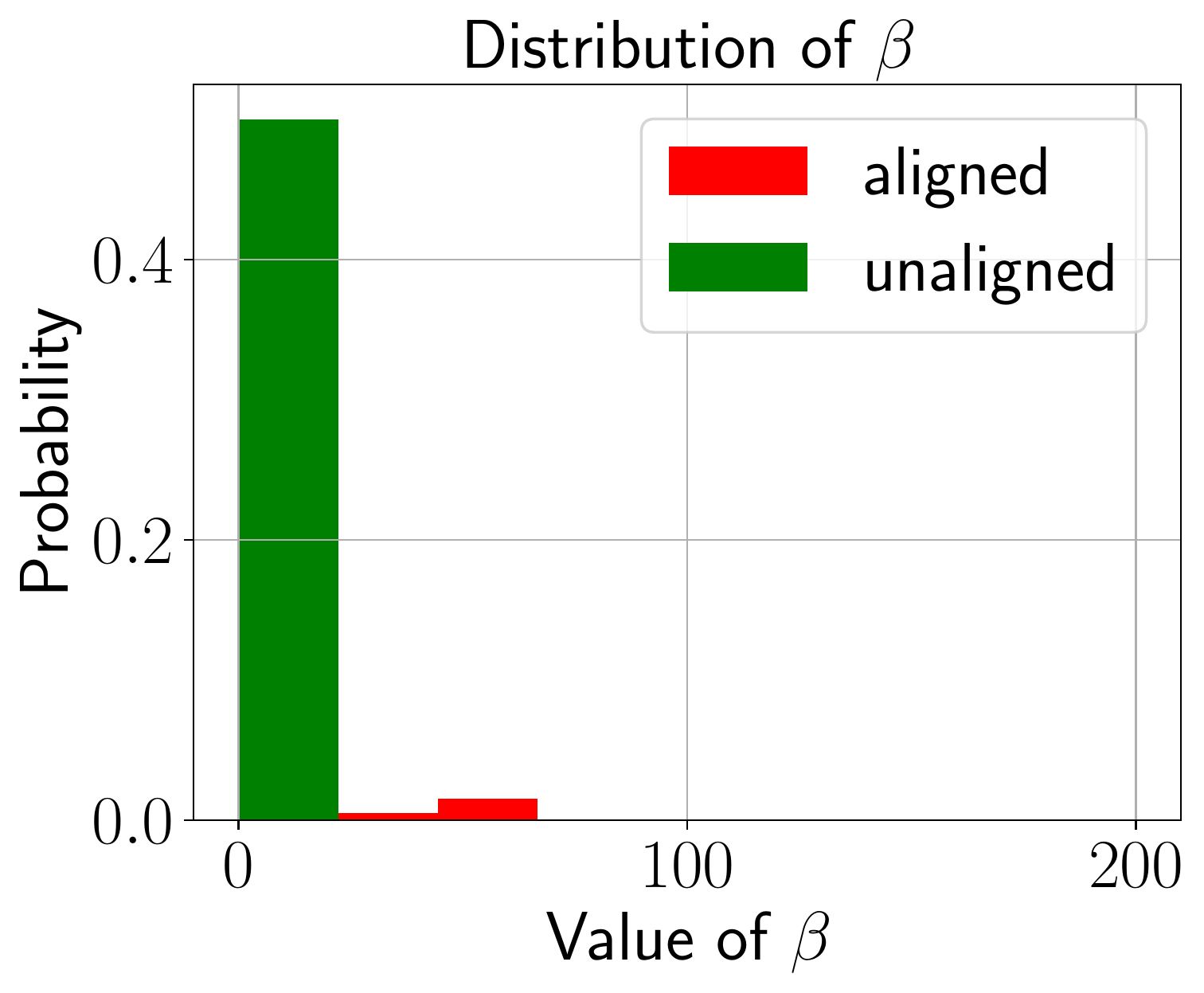}    & \includegraphics[scale=0.2]{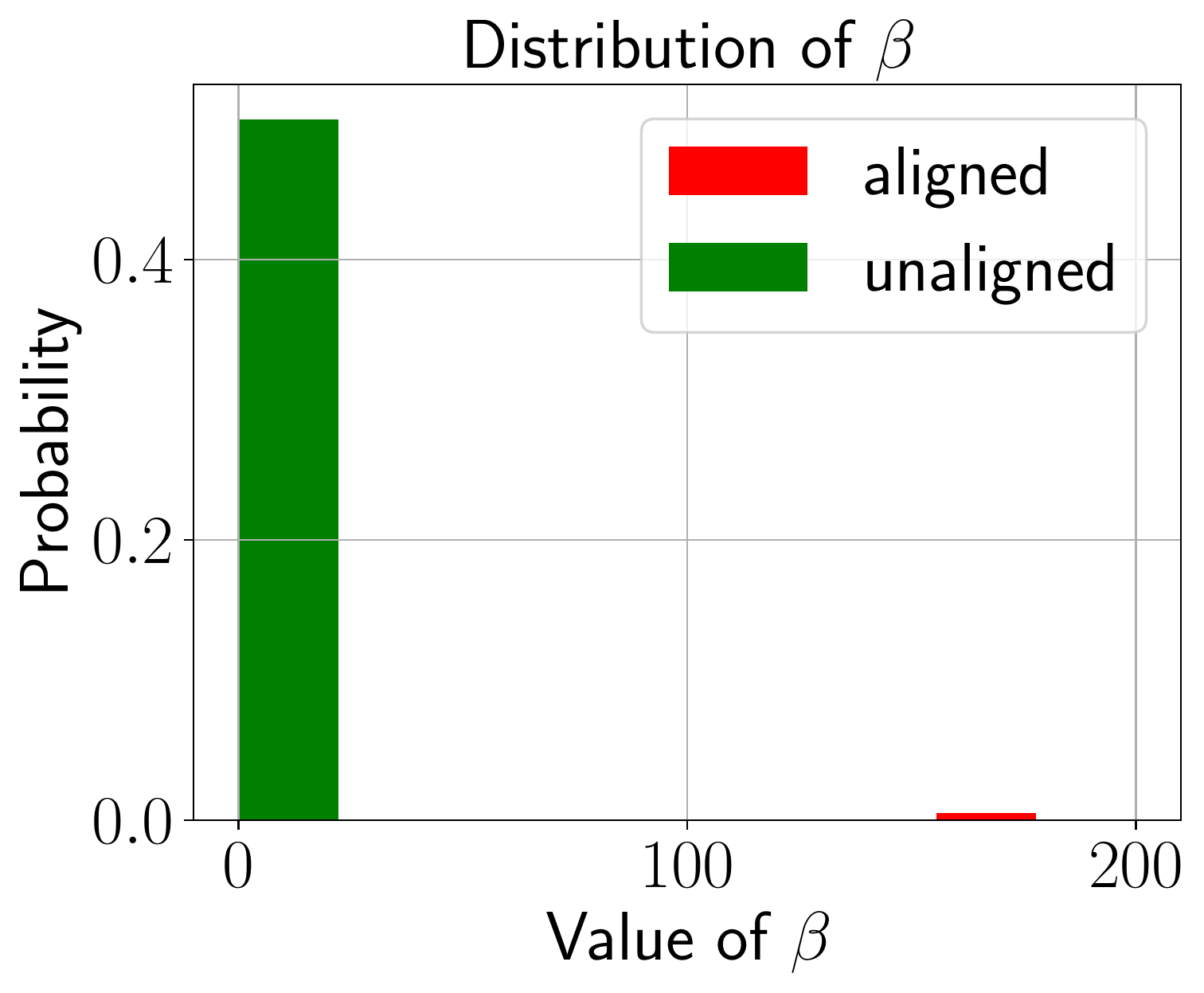}&\includegraphics[scale=0.2]{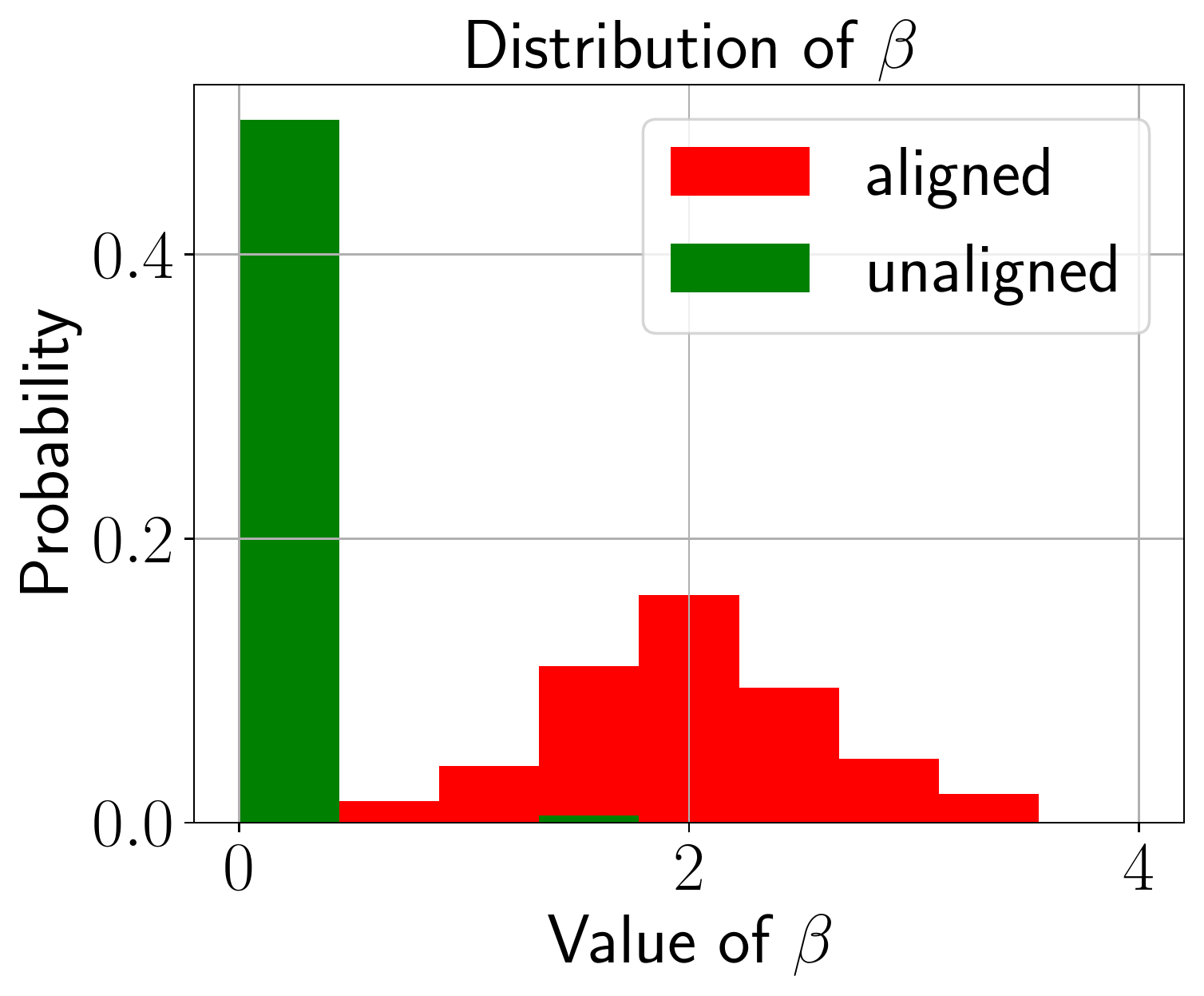}&\includegraphics[scale=0.2]{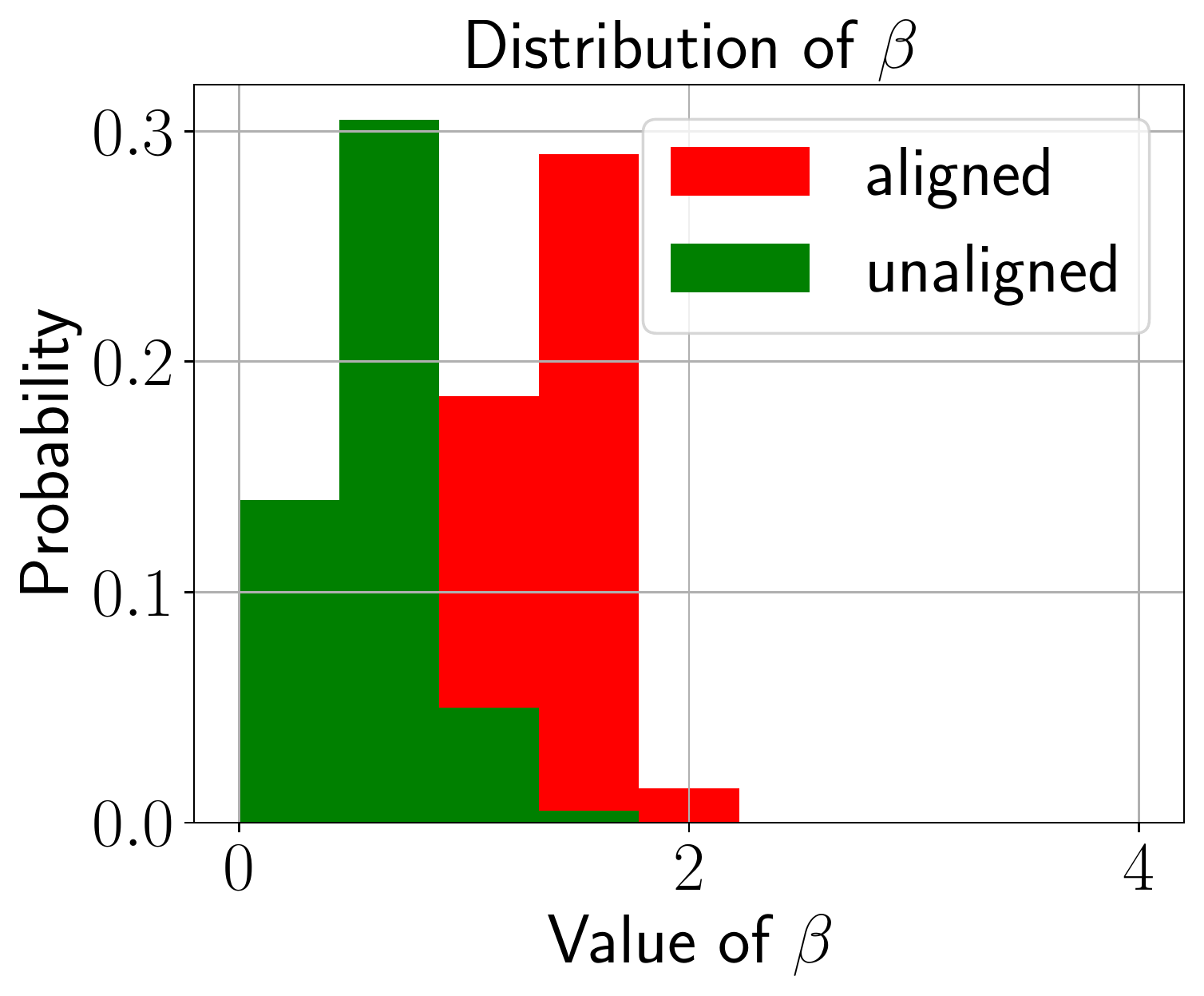} &\includegraphics[scale=0.2]{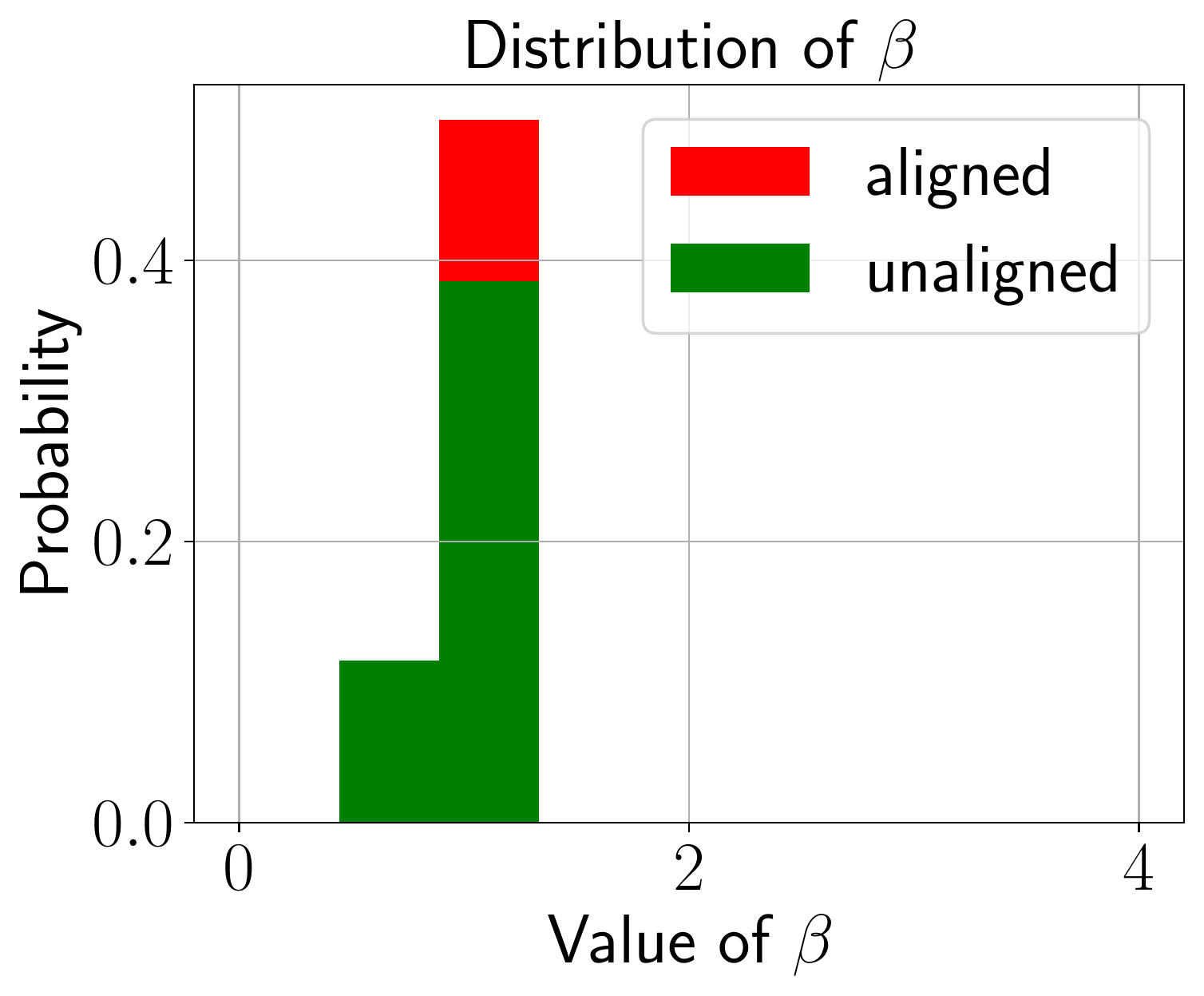}\\
      $\lambda_{\text{ESS}}=0.0$& $\lambda_{\text{ESS}}=0.1$&$\lambda_{\text{ESS}}=1.0$&$\lambda_{\text{ESS}}=3.0$&$\lambda_{\text{ESS}}=10.0$
   \end{tabular} \vspace{-2mm}
    \caption{Distributions of importance weights $\beta_{X}$ with different $\lambda_{\text{ESS}}$ for domain $C_{H}$ in the task $C_{H}\leftrightarrow D_{A}$. We use red to denote importance weights for images in aligned subset $C$ and green for images in unaligned subset $H$.}
    \label{fig:nos_effect}
     \vspace{-0.5cm}
\end{figure*}

\vspace{-0.4cm}
\subsection{Comparisons against Baselines}
\label{subsec:exp_compare}

As can be seen in Figure \ref{fig:comp_syn}, our method can produce good translation results with unaligned domains. In contrast, existing methods are unable to detect the unaligned images and tend to generate  unrelated images. For example, in the task $S \rightarrow A_{U}$, given in the third row, the main task is selfie2anime but most of existing methods are heavily influenced by the Danbooru anime images and hence produce messy results. In particular, Baseline translates a female selfie image to an anime character image rather than the wanted anime face image.

\begin{table}[ht]
    \centering \caption{Precision, recall and accuracy score for the learned $\beta$ in different domains. Baseline denotes our method in which we assign 1 to each sample; its recall is always 1.00.} \vspace{-2mm}
       \begin{small}
    \begin{tabular}{|p{1.2cm}|p{1.4cm}|p{1.4cm}|p{1.4cm}|p{1.4cm}|}
        \hline
         \hfil {Domain} & \hfil {Method} & \hfil {Precision} & \hfil  {Recall} & \hfil  {Accuracy}  \\ \hline
         \hfil \multirow{2}{*}{$H_{S}$}& \hfil Baseline & \hfil 0.55& \hfil \textbf{1.00}& \hfil 0.55\\
          & \hfil IrwGAN & \hfil \textbf{1.00}& \hfil  0.93& \hfil \textbf{0.96}\\\hline
          \hfil\multirow{2}{*}{$Z_{D}$}& \hfil Baseline & \hfil 0.58& \hfil \textbf{1.00}& \hfil 0.58\\
          & \hfil IrwGAN & \hfil \textbf{1.00}& \hfil  0.64& \hfil \textbf{0.79}\\\hline
          \hfil\multirow{2}{*}{$A_{U}$}& \hfil Baseline & \hfil 0.50& \hfil \textbf{1.00}& \hfil 0.50\\
          & \hfil IrwGAN & \hfil \textbf{0.99}& \hfil  {0.97}& \hfil \textbf{0.98}\\ \hline
          \hfil \multirow{2}{*}{$C_{H}$}& \hfil Baseline & \hfil 0.45& \hfil \textbf{1.00}& \hfil 0.45\\
          & \hfil IrwGAN & \hfil \textbf{0.99}& \hfil  0.97& \hfil \textbf{0.98}\\\hline
          \hfil \multirow{2}{*}{$D_{A}$}& \hfil Baseline & \hfil 0.50& \hfil \textbf{1.00}& \hfil 0.50\\
          & \hfil IrwGAN & \hfil \textbf{1.00}& \hfil  \textbf{1.00}& \hfil \textbf{1.00}\\
        \hline
         \hfil \multirow{2}{*}{$T_{B}$}& \hfil Baseline & \hfil 0.50& \hfil \textbf{1.00}& \hfil 0.50\\
          & \hfil IrwGAN & \hfil \textbf{0.80}& \hfil  {0.89}& \hfil \textbf{0.83}\\
        \hline
         \hfil \multirow{2}{*}{$L_{E}$}& \hfil Baseline & \hfil 0.50& \hfil \textbf{1.00}& \hfil 0.50\\
          & \hfil IrwGAN & \hfil \textbf{0.78}& \hfil  {0.85}& \hfil \textbf{0.81}\\\hline
    \end{tabular}
    \end{small}
    \label{tab:beta_acc}
    \vspace{-0.6cm}
\end{table}

Table \ref{tab:exp_fid_kid} reports the FID and KID values of different image translation tasks. Our method outperforms these strong baselines on most datasets. The clear improvement of IrwGAN compared to Baseline method suggests that the importance reweighting scheme plays a crucial role in obtaining good image translation results when domains are unaligned.

\subsection{Analysis of Importance Reweighting}
\label{subsec:exp_irw_loss}
\vspace{-1mm}

Figure \ref{fig:vis_betas} visualizes the learned weights for domain $A_{U}$ in the task $S\rightarrow A_{U}$. As we can see, our method IrwGAN is able to distinguish unaligned images from images in the aligned subsets. Many unwanted anime images, e.g., the full body anime character image in the third row, are assigned very low values of the importance weight. As a consequence, these unwanted images would not affect our image translation process and thus we obtain the best results compared to those by other methods. 

Table \ref{tab:beta_acc} shows the performance the learned $\beta_{X}$ and $\beta_{Y}$ on different unaligned domains. 
For unaligned domain $P_Q$, 
we set labels for images in domain $P$ as 1 and 0 for domain $Q$. 
Since our learned $\beta$ is continuous, for evaluation purposes,  we consider its prediction as 1 if it is above a predefined threshold $0.5$. Our method IrwGAN outperforms the Basline method  by a large margin in terms of precision and accuracy. Note that Baseline achieves a perfect recall score because its prediction is always 1 and thus the false negative is 0.

Importance reweighting helps to recover the aligned subsets from two unaligned domains. It would be interesting to test how many unaligned samples that our method can handle. We first use the existing dataset selfie2anime as the aligned subsets and then add $N$ anime images from the Danbooru dataset to the anime domain. We set $N$ to be $10,30,50,100,150,200,300\%$ of the number of images (3400) in the original anime domain. Results in Table \ref{tab:stress} show that our method is capable of handling the unaligned translation problem under different levels of unaligned samples. We also provide visualizations of learned weights in the supplementary material.

\begin{table}[]
    \centering
     \caption{Results of $S\rightarrow A_U$ under different ratios of unaligned to aligned samples.}
     \vspace{-0.2cm}
    \begin{tabular}{|p{1.4cm}|p{1.6cm}||p{1.4cm}|p{1.6cm}|}
    \hline 
        \hfil  Ratio &\hfil FID $\downarrow$ & \hfil Ratio &\hfil FID $\downarrow$ \\ \hline 
         \hfil 0 $\%$&\hfil 95.25&\hfil 100 $\%$&\hfil 93.73\\ \hline
         \hfil 10 $\%$& \hfil 95.58& \hfil 150 $\%$ &\hfil 92.74\\ \hline
          \hfil 30 $\%$&\hfil 90.54 & \hfil 200 $\%$&\hfil  91.25\\ \hline
         \hfil 50 $\%$& \hfil 93.05&\hfil 300 $\%$ &\hfil 95.44\\ \hline
    \end{tabular}
   \vspace{-0.4cm}
    \label{tab:stress}
\end{table}

\vspace{-1mm}
\subsection{The Effective Sample Size Weight $\lambda_{\text{ESS}}$}
\label{subsec:exp_lambda_nos}
\vspace{-1mm}

$\lambda_{\text{ESS}}$ is designed to control the effective sample size. Figure \ref{fig:nos_effect} shows the distributions of the estimated importance weights $\beta_{X}$ for domain $C_{H}$ with different $\lambda_{\text{ESS}}$ in the task $C_{H}\leftrightarrow D_{A}$. As we can see, if $\lambda_{\text{ESS}}$ is set to 0 or very low, the vector is very sparse, which means our method can only select a few images from the whole domain. As we increase the value of $\lambda_{\text{ESS}}$, the importance weights of more images in unknown aligned subsets are becoming larger while the importance weights of unaligned images are still very small. If we set  $\lambda_{\text{ESS}}$ to a very large value, e.g., 10, the importance weights of all images concentrate around 1.0, which is very close to Baseline that assigns 1 to each sample.

\setlength{\tabcolsep}{1.5pt}
\begin{figure}
    \centering
  \begin{tabular}{cccccc}
      { \footnotesize  Input}& {\footnotesize Ours} &{\footnotesize CycleGAN} &{\footnotesize Input}&{\footnotesize Ours}&{\footnotesize CycleGAN}\\
      \frame{ \includegraphics[width=1.2cm,height=1.2cm]{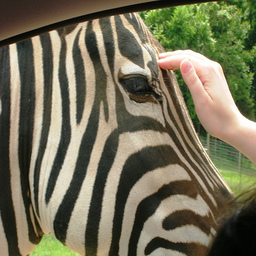}}&
       \frame{\includegraphics[width=1.2cm,height=1.2cm]{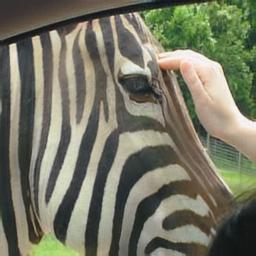}}&
        \frame{ \includegraphics[width=1.2cm,height=1.2cm]{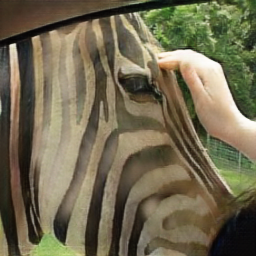}}&
        \frame{ \includegraphics[width=1.2cm,height=1.2cm]{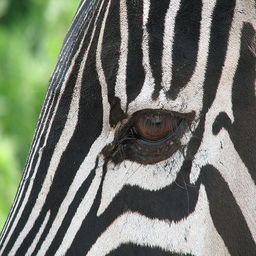}}& \frame{\includegraphics[width=1.2cm,height=1.2cm]{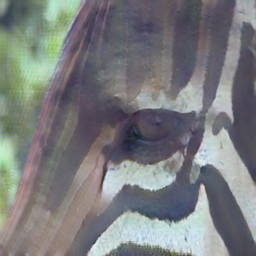}} &
         \frame{ \includegraphics[width=1.2cm,height=1.2cm]{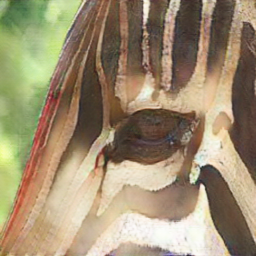}}
        \\
         \multicolumn{6}{c}{ \footnotesize $Z_D\rightarrow H_S$ } \\
          \frame{ \includegraphics[width=1.2cm,height=1.2cm]{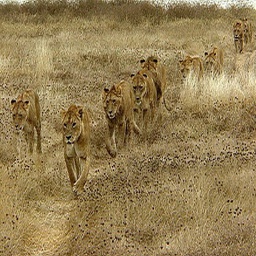}}& \frame{\includegraphics[width=1.2cm,height=1.2cm]{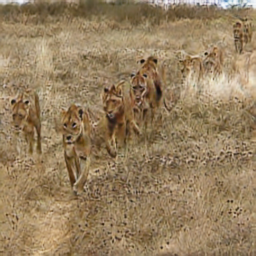}}&
          \frame{ \includegraphics[width=1.2cm,height=1.2cm]{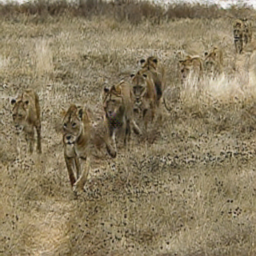}}&
         \frame{ \includegraphics[width=1.2cm,height=1.2cm]{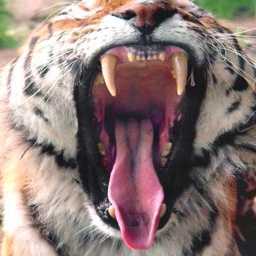}}& \frame{\includegraphics[width=1.2cm,height=1.2cm]{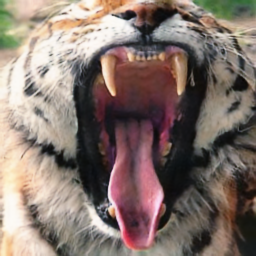}} &
         \frame{ \includegraphics[width=1.2cm,height=1.2cm]{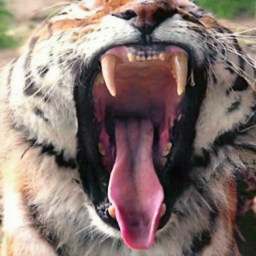}}\\
         \multicolumn{3}{c}{  \footnotesize $L_E \rightarrow T_B$ } &\multicolumn{3}{c}{ $T_B \rightarrow L_E$ }
  \end{tabular}\vspace{-2mm}
    \caption{Some failure cases of IrwGAN.}
    \label{fig:limit}
    \vspace{-0.5cm}
\end{figure}

\vspace{-1mm}
\section{Conclusion and Discussion of Limitations}
\vspace{-1mm}
In this paper, we proposed a novel, more realistic setting for image-to-image translation, in which the two domains are not aligned and hence one has to select suitable images for meaningful translation.  To show that the formulated problem is not only more practical, but also solvable, we developed an importance reweighting-based learning method to automatically select the images and perform translation simultaneously. Our empirical results suggest that it achieves large improvements over existing methods. It is worth noting that our method relies on the assumption that aligned images are easier to be translated to each other. We observe that this hypothesis is generally supported on real images, although it might be violated for some complex images and specific network structure. In other words, it might also be hard to translate some images in the aligned subsets to the other domain. This violation may result in low importance weights on those samples and they will consequently be discarded during training. 


Figure \ref{fig:limit} shows some failure cases of our method. In the first row, we want to translate zebra images to horse images. However, there are few horse head images in the horse domain, which makes it difficult to translate zebra head images to the horse domain. As a consequence, our model will assign low importance to these images and they are discarded during the training.  Our method and CycleGAN both failed on this task. CycleGAN made little changes to the input image while ours method outputs almost identical image to the input.  Similar phenomena happened in the second row. Resolving this issue may require some weak supervision or additional information for representation learning and we leave it as future work. In addition, the domain gap between aligned and unaligned subsets may also be an important factor of the performance. We plan to explore dataset with more diverse domain gaps in the future work.

\textbf{Acknowledgements} We would like to acknowledge the support by the United States Air Force under Contract No. FA8650-17-C-7715, by National Institutes of Health under Contract No. R01HL159805, and by a grant from Apple. The United States Air Force or National Institutes of Health is not responsible for the views reported in this article.
Mingming Gong was supported by Australian Research Council Project DE210101624.

{\small
\bibliographystyle{ieee_fullname}
\bibliography{egbib}
}

\end{document}